\pdfoutput=1
\documentclass[11pt]{article}

\usepackage{emnlp2021}

\usepackage{times}
\usepackage{latexsym}

\usepackage[T1]{fontenc}
\usepackage[utf8]{inputenc}

\usepackage{microtype}

\usepackage{graphicx}
\usepackage{hyperref}

\usepackage[boxruled,linesnumbered,vlined,inoutnumbered]{algorithm2e}
\SetKwInOut{Parameter}{Parameters}
\usepackage{epsfig}
\usepackage{amsmath}
\usepackage{amssymb}
\usepackage{float}
\usepackage{multirow}
\usepackage{hhline}
\usepackage{makecell}
\usepackage{tabularx}

\usepackage{caption}
\usepackage{subcaption}
\usepackage{booktabs}
\usepackage{bbold}
\usepackage{multirow}
\usepackage{array, makecell} %
\usepackage{dsfont}
\usepackage{listings}
\definecolor{coolgrey}{rgb}{0.55, 0.57, 0.67}

\title{Improved Latent Tree Induction with \\Distant Supervision via Span Constraints}

\author{Zhiyang Xu$^{*}$,
  ~Andrew Drozdov\thanks{\hspace{2mm}Equal contribution.}~\hspace{0.5mm},
  ~Jay Yoon Lee,
  ~Tim O'Gorman, \\
  {\bf
  ~Subendhu Rongali,
  ~Dylan Finkbeiner, 
  ~Shilpa Suresh, }
  \\
  {\bf
  ~Mohit Iyyer,
  ~and Andrew McCallum
  } \\
  \AND\\[-4ex]
College of Information and Computer Sciences\\University of Massachusetts Amherst
\AND\\[-4ex]
    \texttt{\{zhiyangxu, adrozdov, jaylee, togorman, srongali,} \\
    \texttt{dfinkbeiner, ssuresh, miyyer, mccallum\}@cs.umass.edu}}

\begin{document}
\maketitle
\begin{abstract}
For over thirty years, researchers have developed and analyzed methods for latent tree induction as an approach for unsupervised syntactic parsing. Nonetheless, modern systems still do not perform well enough compared to their supervised counterparts to have any practical use as structural annotation of text.
In this work, we present a technique that uses distant supervision in the form of span constraints (i.e. phrase bracketing) to improve performance in unsupervised constituency parsing.
Using a relatively small number of span constraints we can substantially improve the output from DIORA, an already competitive unsupervised parsing system. 
Compared with full parse tree annotation, span constraints can be acquired with minimal effort, such as with a lexicon derived from Wikipedia, to find exact text matches.
Our experiments show span constraints based on entities improves constituency parsing on English WSJ Penn Treebank by more than 5 F1. 
Furthermore, our method extends to any domain where span constraints are easily attainable, and as a case study we demonstrate its effectiveness by parsing biomedical text from the CRAFT dataset.
\end{abstract}

\section{Introduction}

\begin{figure}[ht!]
    \centering
    \includegraphics[width=\columnwidth]{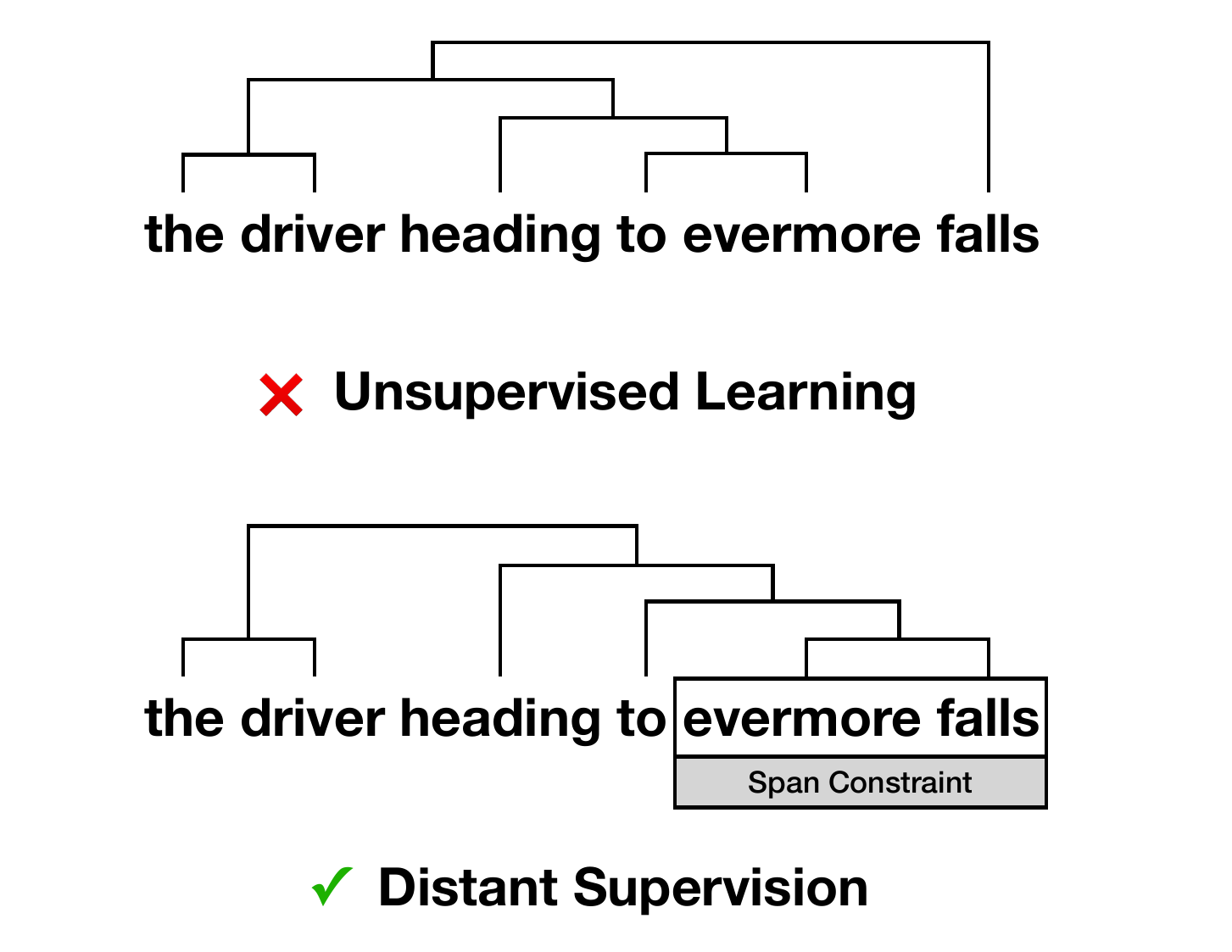}
    \caption{An example sentence and parsing to illustrate distant supervision via span constraints. \emph{Top}: The unsupervised parser predicts a parse tree, but due to natural ambiguity in the text the prediction crosses with a known constraint. \emph{Bottom}: By incorporating the span constraint, the prediction improves and, as a result, recovers the ground truth parse tree. In our experiments, we both inject span constraints directly into parse tree decoding and separately use the constraints only for distant supervision at training time. We find the latter approach is typically more effective.
    }
    \label{fig:first_tree}
\end{figure}
% \textcolor{red}{Alternative sentences: \\ i. (niagara (falls (in (april roars)))) \\ i. ((niagara falls) (in april) roars)}

Syntactic parse trees are helpful for various downstream tasks such as speech recognition \cite{Moore1995CombiningLA}, machine translation \cite{akoury-etal-2019-syntactically}, paraphrase generation \cite{iyyer-etal-2018-adversarial}, semantic parsing \cite{xu-etal-2020-improving}, and information extraction \cite{narad:phdthesis}. While \textit{supervised} syntactic parsers are state-of-the-art models for creating these parse trees, their performance does not transfer well across domains. Moreover, new syntactic annotations are prohibitively expensive; the original Penn Treebank required eight years of annotation \cite{Taylor2003ThePT}, and expanding PTB annotation to a new domain can be a large endeavor. For example, the 20k sentences of biomedical treebanking in the CRAFT corpus required 80 annotator hours per week for 2.5 years, include 6 months for annotator training \cite{Verspoor2011ACO}. However, although many domains and many languages lack full treebanks, they do often have access to other annotated resources such as NER, whose spans might provide some partial syntactic supervision. We explore whether \textit{unsupervised} parsing methods can be enhanced with distant supervision from such spans to enable the types of benefits afforded by supervised syntactic parsers without the need for expensive syntactic annotations.

% -- it's estimated this amount of collection requires 80 hours per week commitment for 2.5 years, which includes 6 months for annotator training \cite{Verspoor2011ACO}. Recent interest in \textit{unsupervised} parsing as a research topic is in part to develop systems that can automatically infer constituent-like structures with limited, if any, supervision. These systems will hopefully enable the types of benefits afforded by supervised syntactic parsers at a decreased cost and larger scale.

% version 1
% We aim to ``bridge the gap'' between supervised and unsupervised parsing with distant supervision through span constraints. These span constraints indicate that a certain sequence of words in a sentence form a constituent span in its parse tree, and we obtain these partial ground-truths without explicit user annotation. Using small amounts of distant supervision has been helpful in both supervised \cite{Finkel2009JointPA,Ganchev2010PosteriorRF} and unsupervised \cite{haghighi-klein-2006-prototype,smith-eisner-2006-annealing,cao-etal-2020-unsupervised} syntactic parsing. We take inspiration from these works, and design a novel fully neural system that improves a competitive neural unsupervised parser (DIORA; \citealt{Diora}) using span constraints defined on a portion of the training data. In the large majority of cases, the number of spans constraints per sentence is much lower than that specified by a full parse tree. We find that entity spans are effective as constraints, and can readily be acquired from existing data or derived from a gazetteer.

We aim to ``bridge the gap'' between supervised and unsupervised parsing with distant supervision through span constraints. These span constraints indicate that a certain sequence of words in a sentence form a constituent span in its parse tree, and we obtain these partial ground-truths without explicit user annotation. We take inspiration from previous work incorporating distant supervision into parsing \cite{haghighi-klein-2006-prototype,Finkel2009JointPA,Ganchev2010PosteriorRF,cao-etal-2020-unsupervised}, and design a novel fully neural system that improves a competitive neural unsupervised parser (DIORA; \citealt{Diora}) using span constraints defined on a portion of the training data. In the large majority of cases, the number of spans constraints per sentence is much lower than that specified by a full parse tree. We find that entity spans are effective as constraints, and can readily be acquired from existing data or derived from a gazetteer.

% Using small amounts of supervisio\usepackage{}n has previously been helpful in unsupervised parsing. \citet{cao-etal-2020-unsupervised} use a description of linguistic transformations that are used to create \textit{constituency tests}. \citet{haghighi-klein-2006-prototype} use a list of part-of-speech \textit{prototypes} for labeled parsing --- any phrase that has the same part-of-speech as one of the templates is assigned a specific label.

% In contrast, we explore the effectiveness of a very general type of constraint. All that is required are spans that would reasonably appear as a constituent. Specifically, we find that entity spans are effective as constraints, and can readily be acquired from existing data or derived from a gazetteer.

%In our work, we take an approach of improving an unsupervised parser by injecting span constraints as a source of distant supervision. Specifically, 

% Partially Structured SVM (PS-SVM) because it resembles the Structured SVM loss often used in supervised constituency parsing \cite{stern-etal-2017-minimal,kitaev-klein-2018-constituency} but only requires that a few spans be provided as constraints rather than a full parse tree.

In our experiments, we use DIORA as our baseline and improve upon it by injecting these span constraints as a source of distant supervision. We introduce a new method for training DIORA that leverages the structured SVM loss often used in supervised constituency parsing \cite{stern-etal-2017-minimal,kitaev-klein-2018-constituency}, but only depends on partial structure. We refer to this method as partially structured SVM (PS-SVM). Our experiments indicate PS-SVM improves upon unsupervised parsing performance as the model adjusts its prediction to incorporate span constraints (depicted in Figure \ref{fig:first_tree}). Using ground-truth entities from Ontonotes \cite{Conll2012} as constraints, we achieve more than 5 F1 improvement over DIORA when parsing English WSJ Penn Treebank \cite{marcus1993building}. Using automatically extracted span constraints from an entity-based lexicon (i.e. gazetteer) is an easy alternative to ground truth annotation and gives 2 F1 improvement over DIORA. Importantly, training DIORA with PS-SVM is more effective than simply injecting available constraints into parse tree decoding at test time. We also conduct experiments with different types of span constraints. Our detailed analysis shows that entity-based constraints are similarly useful as the same number of ground truth NP constituent constraints. Finally, we show that DIORA and PS-SVM are helpful for parsing biomedical text, a domain where full parse tree annotation is particularly expensive.

% TODO: We present the first neural model for unsupervised parsing that learns tree structure and incorporates spans constraints, without being completely reliant on said constraints.

% In our work, we aim to improve the parsing output from DIORA \cite{Diora} because it is a competitive unsupervised parser amenable to distant supervision using span constraints. DIORA is a chart parser, and trees are extracted using CKY algorithm. In our case, we constrain CKY to include the span constraints, and then supervise the model to output the constrained tree.

%% contributions.

%% Type of analysis provided.

% Third paragraph, % Applying distant supervision.

% \begin{itemize}
%     \item We use DIORA because it is a high performing parser amenable to incorporating partial supervision on trees through constrained CKY.
%     \item It is not enough to gather constraints: 1) The constraints might not agree with the downstream task, so you want a system with soft constraints; 2) Constrained CKY is not effective because it is too different than training, and only applies to sentences with constraints.
%     \item We start with pre-trained model.
%     \item End with main contributions and results.
% \end{itemize}

\section{Background: DIORA}

The Deep Inside-Outside Recursive Autoencoder (DIORA; \citealp{Diora}) is an extension of tree recursive neural networks (TreeRNN) that does not require pre-defined tree structure. It depends on the two primitives $\mathrm{Compose} : \mathbb{R}^{2D} \rightarrow \mathbb{R}^{D}$ and $\mathrm{Score} : \mathbb{R}^{2D} \rightarrow \mathbb{R}^{1}$. DIORA is bi-directional --- the \textit{inside pass} builds phrase vectors and the \textit{outside pass} builds context vectors. DIORA is trained by predicting words from their context vectors, and has been effective as an unsupervised parser by extracting parse trees from the values computed during the inside pass.

\paragraph{Inside-Outside} Typically, a TreeRNN would follow a parse tree to continually compose words or phrases until the entire sentence is represented as a vector, but this requires knowing the tree structure or using some trivial structure such as a balanced binary tree. Instead of using a single structure, DIORA encodes all possible binary trees using a soft-weighting determined by the output of the score function. There are a combinatorial number of valid parse trees for a given sentence --- it would be infeasible to encode each of them separately. Instead, DIORA decomposes the problem of representing all valid parse trees by encoding all subtrees over a span into a single phrase vector. For example, each phrase vector is computed in the inside pass according to the following equations:

\begin{align*}
    h^{in}_{i,j} &= \sum^{j-1}_{k=i+1} s^{in}_{i,j,k} h^{in}_{i,j,k} \\
    h^{in}_{i,j,k} &= \mathrm{Compose}(h^{in}_{i,k}, h^{in}_{k,j}) \\
    s^{in}_{i,j,k} &= \mathrm{Score}(h^{in}_{i,k}, h^{in}_{k,j}) + s^{in}_{i,k} + s^{in}_{k,j} \\
\end{align*}

% TODO: Include outside pass and reconstruction loss.

\noindent
The outside pass is computed in a similar way:

\begin{align*}
    h^{out}_{i,j} =& \sum^{j-1}_{k=i+1} s^{out}_{i,j,k} h^{out}_{i,j,k} \\
    h^{out}_{j,k} =& 
        \mathbb{1}_{j<k} \cdot ~\mathrm{Compose}(h^{in}_{j,k}, h^{out}_{i,k}) ~+ \\ 
    &   \mathbb{1}_{j<k} \cdot  ~\mathrm{Compose}(h^{in}_{k,i}, h^{out}_{k,j}) \\
    s^{out}_{i,j,k} =& 
    \mathbb{1}_{j<k} \cdot \big[ ~\mathrm{Score}(h^{in}_{j,k}, h^{out}_{i,k}) + s^{in}_{j,k} + s^{out}_{i,k} \big] \\
&   \mathbb{1}_{j<k} \cdot \big[ ~\mathrm{Score}(h^{in}_{k,i}, h^{out}_{k,j}) + s^{in}_{k,i} + s^{out}_{k,j} \big] ~\text{,}
\end{align*}

\noindent
where $\mathbb{1}_{j<k}$ is an indicator function that is $1$ when the sibling span is on the right, and $0$ otherwise (see Figure 2 in \citealp{drozdov2020sdiora} for a helpful visualization of the inside and outside pass).

\paragraph{Training} DIORA is trained end-to-end directly from raw text and without any parse tree supervision. In our work, we use the same reconstruction objective as in \citet{Diora}. For a sentence $x$, we optimize the probability of the $i$-th word $x_i$ using its context (${x}_{-i}$):

\begin{align*}
    J_{rec} &= -\frac{1}{|x|} \sum_{i = 0}^{|x| - 1} \log P(x_i | x_{-i}) ~\text{,}
\end{align*}

\noindent
where $P(.)$ is computed use a softmax layer over a fixed vocab with the outside vector ($h^{out}_{i,i}$) as input.

\paragraph{Parsing} DIORA has primarily been used as an unsupervised parser. This requires defining a new primitive $\mathrm{TreeScore} : S(y) = \sum_{i,j,k \in y} s^{in}_{i,j,k}$. A tree $y$ can be extracted from DIORA by solving the search problem that can be done efficiently with the CKY algorithm \cite{CKY1,CKY2}: 

\begin{align*}
    CKY(x) &= \arg\max_{y} S(y)
\end{align*}

\section{Injecting Span Constraints to DIORA}

% \jys{Let's rewrite this paragarpah.}\prev{If there is available data that provides a mapping from sentences $X$ to constituency trees $Y$, then the most effective way to predict annotations for new sentences is to train a supervised parser. In some cases, it is expensive to acquire data that contains $Y$, but attaining a set of spans that presumably would be contained in $Y$ is more feasible.} 

In this section, we present a method to improve parsing performance by training DIORA such that trees extracted through CKY are more likely to contain known span constraints.

% we present a method to train DIORA such that the trees extracted through CKY will contain known span constraints.

\subsection{Test-time injection: Constrained CKY}

\label{sec:ccky_weaknesses}

% First, we define a modified search problem that can be solved with a constrained version of CKY: 
% \jy{ First, we define a constrained CKY (CCKY) that modifies the search problem to incorporate span constraints:}

% \begin{align*}
%     CCKY(X, z) =& \arg\max_{y} \big[ S(y) + \epsilon \cdot g(y, z) \big],
% \end{align*}

% \noindent
% where $Z$ is a set of known span constraints for $X$, \jy{$g(Y, Z)$ measures how well the span constraints $Z$ are satisfied in $Y$, i.e. } $g(Y, Z) = \sum_{z \in Z} \mathbb{1}(z \in Y)$, and $\epsilon$ \prev{ is a positive variable that guarantees} \jy{is an importance weight for the span constraint to guarantee} the highest scoring trees are the ones that satisfy the most constraints.

One option to improve upon CKY is to simply find span constraints and then use a constrained version of CKY (CCKY):

\begin{align*}
    CCKY(x, z) =& \arg\max_{y} \big[ S(y) + \epsilon \cdot g(y, z) \big],
\end{align*}

\noindent
where $z$ is a set of known span constraints for $x$, $g(y, z)$ measures how well the span constraints are satisfied in $y$, i.e.  $g(y, z) = \sum_{i=0}^{|z| - 1} \mathbb{1}(z_i \in y)$, and $\epsilon$ is an importance weight for the span constraint to guarantee the highest scoring trees are the ones that satisfy the most constraints.\footnote{To save space, we exclude $\epsilon$ hereafter.} Using CCKY rather than CKY typically gives a small boost to parsing performance, but has several downsides described in the remainder of this subsection.

\paragraph{Can overfit to constraints} DIORA learns to assign weights to the trees that are most helpful for word prediction. For this reason, it is logical to use the weights to find the highest scoring tree. With $\textsc{CCKY}$, we can find the highest scoring tree that also satisfies the constraints, but this tree could be very different from the original output. Ideally, we would like a method that can incorporate span constraints in a productive way that is not detrimental to the rest of the structure.
% \footnote{An alternative to constrained CKY would be to find the tree with minimal edit distance with the original output. Similarly, there could be some benefit by looking for the tree satisfying the constraint and getting highest F1 compared with the original tree.}

\paragraph{Only benefits sentences with constraints} If we are dependent on constraints for CCKY, then only sentences that have said constraints will receive any benefit. Ideally, we would like an approach where even sentences without constraints could receive some improvement.

\paragraph{Constraints are required at test time} If we are dependent on constraints for CCKY, then we need to find constraints for every sentence at test time. Ideally, we would like an approach where constraints are only needed at the time of training.

% We aim to find constraints that align with constituency structure, but 
\paragraph{Noisy constraints} Occasionally a constraint disagrees with a comparable constituency parse tree. In these cases, we would like to have an approach where the model can choose to include only the most beneficial constraints.

\begin{table*}[!htbp]
\begin{center}
\resizebox{\linewidth}{!}{%
\begin{tabular}{l | l | l | l}
\toprule
Loss & $\alpha$ & $y^-$ & $y^+$ \\
\midrule
$\textsc{NCBL}$ & $1$ & $\arg\max_{y} S(y)$ & $\arg\max_{y} [ S(y) + g(y,z) ] $ \\
$\textsc{Min Difference}$ & $1$ & $\arg\max_{y} S(y)$ & $\arg\max_{y} [ S(y) + g(y, z)  + g(y, y^-) ] $  \\
$\textsc{Rescale}$ & $g(y^+, y^-)$ & $\arg\max_{y} S(y)$ & $\arg\max_{y} [ S(y) + g(y,z) ] $ \\
$\textsc{Structured Ramp}$ & $1$ & $\arg\max_{y} [ S(y) - g(y, z) ] $  & $\arg\max_{y} [ S(y) + g(y,z) ] $ \\
\bottomrule
\end{tabular}
}
\end{center}
\vspace{-0.4cm}
\caption{Multiple variants of the Partially Structured SVM (PS-SVM) loss, $J_{PS} = \alpha \cdot \max(0, 1 + S(y^-) - S(y^+))$, where $z$ denotes constraint spans and $g(y, z) = \sum_{i=0}^{|z| - 1} \mathbb{1}(z_i \in y)$.}
% In our experiments we encourage different parsing behavior by separately applying 4 variants of the Partially Structured SVM (PS-SVM) loss, $J_{PS} = \alpha \cdot \max(0, 1 + S(Y) - S(Y^*))$. Constrained CKY: Uses the highest scoring tree as negative example, and constrained highest scoring tree as positive. Min Difference: Uses the highest scoring tree that both satisfies the constraints and makes minimal number of errors with respect to negative tree. Rescale: Similar to Min Difference, except scales loss proportional to number of errors. Structured Ramp: The negative tree should not satisfy any constraints.
\label{tab:loss_eq}
\end{table*}
% TODO: Make note where we learned about structured ramp.

% START: PS-SVM
% \subsection{Partially Structured SVM}

% \begin{align}
%     J &= \alpha \cdot \max(0, 1 + S(Y^-) - S(Y^+)) \nonumber
% \end{align}

% \subsection*{Constrained CKY}

% \begin{align*}
%     Y^+ &= \arg\max_{y} [ S(y) - \Delta(z, y) ] 
% \end{align*}

% \subsection*{Constrained CKY (Min Difference)}

% \begin{align*}
%     Y^- &= \arg\max_{y} S(y) \\
%     Y^+ &= \arg\max_{y} [ S(y) - \Delta(Y^-, y) ] 
% \end{align*}

% \subsection*{Constrained CKY (Rescale)}

% \begin{align*}
%     \alpha &= F1(Y^-, Y^+)
% \end{align*}

% \subsection*{Structured Ramp Loss}

% \begin{align*}
%     Y^- &= \arg\max_{y} [ S(y) + \Delta(z, y) ]  \\
%     Y^+ &= \arg\max_{y} [ S(y) - \Delta(z, y) ] 
% \end{align*}

% END: PS-SVM

\subsection{\parbox[t]{15em}{Distant Supervision: \\ Partially Structured SVM}}

\vspace{2mm}

To address the weaknesses of $\textsc{CCKY}$ we present a new training method for DIORA called Partially Structured SVM (PS-SVM).\footnote{PS-SVM can be loosely thought of as an application-specific instantiation of Structural SVM with Latent Variables \cite{Yu2009LearningSS}.} This is a training objective that can incorporate constraints during training to improve parsing and addresses the aforementioned weaknesses of constrained CKY. PS-SVM follows these steps:

% , and similar approaches have been done for other tasks such as text segmentation and co-reference resolution.

\begin{enumerate}
    \item Find a \textit{negative} tree ($y^-$), such as the highest scoring tree predicted by the model: \\ $y^- \leftarrow CKY(x)$.
    \item Find a \textit{positive} tree ($y^+$), such as the highest scoring tree that satisfies known constraints: \\
    $y^+ \leftarrow CCKY(x, z)$.
    \item Use the structured SVM with fixed margin to learn to include constraints in the output: \\ $J_{PS} = \alpha \cdot \max(0, 1 + S(y^-) - S(y^+))$.
\end{enumerate}

\newpage

\subsection{Variants of Partially Structured SVM \label{sec:ps_variants}}

The most straightforward application of PS-SVM assigns $y^+$ to be the highest scoring tree that also incorporates known constraints, and we call this \textbf{$\textsc{Naive Constraint-based Learning}$ ($\textsc{NCBL}$)}. The shortcoming of $\textsc{NCBL}$ are similar to $\textsc{CCKY}$, $y^+$ may be drastically different from the initial prediction $y^-$ and the model may \textit{overfit} to the constraints. With this in mind, an alternative to $\textsc{NCBL}$ is to find $y^+$ that is high scoring, satisfies the constraints, and has the minimal number of differences with respect to $y^-$. We refer to this approach as \textbf{$\textsc{Min Difference}$}.

The $\textsc{Min Difference}$ approach gives substantial weight to the initial prediction $y^-$, which may be helpful for avoiding overfitting to the constraints, but simultaneously is very restrictive on the region of positive trees. In other constraint-based objectives for structured prediction, such as gradient-based inference \cite{GradientBased}, the agreement with constraints is incorporated as a scaling penalty to the gradient step size rather than explicitly restricting the search space of positive examples. Inspired by this, we define another alternative to $\textsc{NCBL}$ called \textbf{$\textsc{Rescale}$} that scales the step size based on the difference between $y^+$ and $y^-$. If the structures are very different, then only use a small step size in order to both prevent overfitting to the constraints and allow for sufficient exploration.

For margin-based learning, for stable optimization a technique known as loss-augmented inference assigns $y^-$ to the be the highest scoring and most offending example with respect to the ground truth. When a full structure is not available to assign $y^+$, then an alternative option is to use the highest scoring prediction that satisfies the provided partial structure. This approach is called \textbf{$\textsc{Structured Ramp}$} loss \cite{do_structured_ramp,gimpel-smith-2012-structured,shi2021learning}.

% In Table \ref{tab:loss_eq} we define the $\textsc{CCKY}$, $\textsc{MinDifference}$, $\textsc{Rescale}$, and $\textsc{Structured Ramp}$ variants of PS-SVM.

In Table \ref{tab:loss_eq} we define the 4 variants of PS-SVM. Variants that do not use loss-augmented inference have gradient 0 when $y^-$ contains all constraints.

\section{Experimental Setup}

% version a :With our experiments we aim to answer specific research questions about unsupervised parsing augmented with span constraints:
% The main motivation of our experiments is to prove the span constraints as distant supervision can benefit the unsupervised parsing. With our experiments we aim to answer the following research questions:   
% \begin{enumerate}
%     % Is a relatively small set of span constraints sufficient for improving unsupervised parsing
%     \item[a.] The number of span constraints derived from ground truth data is small comparing to the total number of spans in the parse tree. Are those limited supervision signal from span constraints sufficient for improving unsupervised parsing?
%     % Which types of constraints are most effective?
%     \item[b.] There are different types of constraints based on their syntactic roles in the parse tree. Diora does a decent job on capturing some of them while missing a substantial amount on others. In our experiments we show that some of the constraints are inherently more informative than others and more effective in guiding the model.
%     \item[c.] Is it feasible to collect enough span constraints for our approach?
%     \item[d.] Can the span constraints push other incorrect spans to become correct?
% \end{enumerate}

In this section, we provide details on data pre-processing, running experiments, and evaluating model predictions. In addition, code to reproduce our experiments and the model checkpoints are available on Github.\footnote{\href{https://github.com/iesl/distantly-supervised-diora}{https://github.com/iesl/distantly-supervised-diora}}

\subsection{Training Data and Pre-processing}

We train our system in various settings to verify the effectiveness of PS-SVM with span constraints. In all cases, we require access to a text corpus with span constraints.\footnote{Appendix \ref{sec:app_constraint} provides further details about constraints.}
\paragraph{Ontonotes} (CoNLL 2012; \citealt{Conll2012}) consists of ground truth named entity and constituency parse tree labels. In our main experiment (see Table \ref{tab:parsing_entities}), we use the $57,757$ ground truth entities from training data as span constraints. 

% \vfill
% \noindent
\paragraph{WSJ Penn Treebank} \cite{marcus1993building} consists of ground truth constituency parse tree labels. It is an often-used benchmark for both supervised and unsupervised constituency parsing in English. We also derive synthetic constraints using the ground truth constituents from this data.

% \vfill
% \noindent
\paragraph{MedMentions} \cite{Mohan2019MedMentionsAL} is a collections of Pubmed abstracts that have been annotated with UMLS concepts. This is helpful as training data for the biomedical domain. For training we only use the raw text to assist with domain adaptation. We tokenize the text using scispacy.

% \vfill
% \noindent
\paragraph{The Colorado Richly Annotated Full Text (CRAFT)} \cite{Cohen2017TheCR} consists of biomedical journal articles that have been annotated with both entity and constituency parse labels. We use CRAFT both for training (with $18,448$ entity spans) and evaluation of our model's performance in the biomedical domain. We sample 3k sentences of training data to use for validation.

% Below we list the number of sentences in each dataset. TODO: Make this a table.

% \begin{itemize}
%     \item Ontonotes
%     \item WSJ Penn Treebank
%     \item MedMentons
%     \item CRAFT
% \end{itemize}

\subsubsection{Automatically extracted constraints}

We experiment with two settings where span constraints are automatically extracted from the training corpus using dictionary lookup in a lexicon. These settings simulate a real world setting where full parse tree annotation is not available, but partial span constraints are readily available.

\paragraph{PMI Constraints} We use the phrases defined in the vocab from \citet{Mikolov2013DistributedRO} as a lexicon, treating exact matches found in Ontonotes as constraints. The phrases are learned through word statistics by applying pointwise mutual information (PMI) to find relevant bi-grams, then replacing these bi-grams with a new special token representing the phrase --- applied multiple times this technique is used to find arbitrarily long phrases.

% \section*{Part-of-Speech Templates}

% We use ground truth constituents to find part-of-speech patterns that frequently appear as phrases. We use the most frequent patterns to define templates that can be used to extract phrases in arbitrary text. 

\paragraph{Gazetteer} We use a list of 1.5 million entity names automatically extracted from Wikipedia \cite{ratinov-roth-2009-design}, which has been effective for supervised entity-centric tasks with both log-linear and neural models \cite{liu-etal-2019-towards}. We derive constraints by finding exact matches in the Ontonotes corpus that are in the gazetteer. A lexicon containing entity names is often called a gazetteer.

\subsection{Training Details}

In all cases, we initialize our model's parameters from pre-trained DIORA \cite{Diora}. We then continue training using a combination of the reconstruction and PS-SVM loss. Given sentence $x$ and constraints $z$, the instance loss is:

\begin{align*}
    J(x, z) = J_{rec}(x) + J_{PS}(x, z)
\end{align*}

For the newswire domain, we train for a maximum of 40 epochs on Ontonotes using 6 random seeds and use grid search, taking the best model in each setting according to parsing F1 on the PTB validation set. For biomedical text, since it is a shift in domain from the DIORA pre-training, we first train for 20 epochs using a concatenation of MedMentions and CRAFT data with only the reconstruction loss\footnote{The training jointly with MedMentions and CRAFT is a special case of ``intermediate fine-tuning'' \cite{Phang2018SentenceEO}.} (called DIORA$_{ft}$ for ``fine-tune''). Then, we train for 40 epochs like previously mentioned, using performance on a subset of 3k random sentences from the CRAFT training data for early stopping. Hyperparameters are  in Appendix \ref{sec:app_hyperparam}.

% 

% TODO: \footnote{\citet{Shi_2019} point out that validation sets can disproportionally skew performance of unsupervised parsing systems. We re-did early stopping using 100 random sentences and found that the best model remained the same in all cases. This is consistent with the DIORA-related experiments in \citet{Shi_2019} that show DIORA performance is robust when only a small number of samples are used for model selection.}

% TODO: Fill this space.

\newpage

% Includes mean and stdev
\begin{table}[ht!]
\setlength\tabcolsep{4pt}
\begin{center}
\begin{tabular}{l | l  }
\toprule
 & \multicolumn{1}{c}{F1}  \\
\midrule
{\small \bf General Purpose} &  \\
Ordered Neuron$\dagger$  \cite{shen2018ordered}     & 48.1 \scriptsize${\pm  1.0}$ \\
Compound PCFG$\dagger$  \cite{kim2019compound}      & 55.2 \scriptsize${\pm  2.5}$ \\
DIORA$\ddagger$ \cite{Diora}                    & 56.8  \\ 
S-DIORA$\dagger$  \cite{drozdov2020sdiora}          & 57.6 \scriptsize${\pm  3.2}$  \\
\midrule
{\small \bf Constituency Tests} &   \\
RoBERTa$\dagger$  \cite{cao-etal-2020-unsupervised} & \textbf{62.8} \scriptsize${\pm  1.6}$  \\
\midrule
{\small \bf DIORA Span Constraints} &  \\
+${\textsc{CCKY}}$                           & 57.5 \\
+${\textsc{PS-SVM}_\textsc{NCBL}}$           & 60.4 \scriptsize${\pm  0.1}$    \\
+${\textsc{PS-SVM}_\textsc{Mindiff}}$        & 59.0 \scriptsize${\pm  0.8}$    \\
+${\textsc{PS-SVM}_\textsc{Rescale}}$        & \textit{61.2} \scriptsize${\pm  0.6}$   \\
+${\textsc{PS-SVM}_\textsc{Structure Ramp}}$ & 59.9 \scriptsize${\pm  1.0}$     \\
\bottomrule
\end{tabular}
\end{center}
\vspace{-0.4cm}
\caption{Parsing F1 on PTB. The average F1 across random seeds is measured on the test set, and the standard deviation is shown as subscript when applicable. $\dagger$: Indicates that standard deviation is the approximate lower bound derived from the mean, max, and number of random seeds. $\ddagger$: Indicates no average performance available, so the max is reported.}
\label{tab:parsing_entities}
\end{table}

\subsection{Evaluation}

In all cases, we report Parsing F1 aggregated at the sentence level --- F1 is computed separately for each sentence then averaged across the dataset. To be consistent with prior work, punctuation is removed prior to evaluation\footnote{In general, it is less important that subtrees associated with punctuation match the Penn Treebank guideline \cite{bies1995bracketing} than if the model makes consistent decision with respect to these cases. For this reason, omitting punctuation for evaluation gives a more reliable judgement when parsing is unsupervised.} and F1 is computed using the eval script provided by \citet{shen2017neural}.\footnote{This script ignores trivial spans, and we use the version provided in \href{https://github.com/harvardnlp/compound-pcfg}{https://github.com/harvardnlp/compound-pcfg}.}\textsuperscript{,}\footnote{We were not able to reproduce the results from the concurrent work \citet{shi2021learning}, which does not share their parse tree output and uses a slightly different evaluation.} In tables \ref{tab:parsing_entities}, \ref{tab:real_world}, and \ref{tab:craft} we average performance across random seeds and report the standard deviation.

\paragraph{Baselines} In Table \ref{tab:parsing_entities}, we compare parsing F1 with four \textit{general purpose} unsupervised parsing models that are trained directly from raw text. We also compare with \citet{cao-etal-2020-unsupervised} that uses a small amount of supervision to generate constituency tests used for training --- their model has substantially more parameters than our other baselines and is based on RoBERTa \cite{Liu2019RoBERTaAR}.

\newpage

\section{Results and Discussion}

In our experiments and analysis we aim to address several research questions about incorporating span constraints for the task of unsupervised parsing.

\subsection{Is Constrained CKY sufficient? }

A natural idea is to constrain the output of DIORA to contain any span constraints (\S\ref{sec:ccky_weaknesses}). We expect this type of \textit{hard constraint} to be ineffective for various reasons: 1) The model is not trained to include constraints, so any predictions that forces their inclusion are inherently noisy; 2) Similar to (1), some constraints are not informative and may be in disagreement with the desired downstream task and the model's reconstruction loss; and 3) Constraints are required at test time and only sentences with constraints can benefit.

We address these weaknesses by training our model to include the span constraints in its output using PS-SVM. This can be considered a \textit{soft} way to include the constraints, but has other benefits including the following: 1) The model implicitly learns to ignore constraints that are not useful; 2) Constraints are not necessary at test time; and 3) The model improves performance even on sentences that did not have constraints.

The effectiveness of our approach is visible in Table \ref{tab:parsing_entities} where we use ground truth entity boundaries as constraints. CCKY slightly improves upon DIORA, but our PS-SVM approach has a more substantial impact. We experiment with four variants of PS-SVM (described in \S\ref{sec:ps_variants}) --- $\textsc{Rescale}$ is most effective, and throughout this text this is the variant of PS-SVM used unless otherwise specified.

\subsection{Real world example with low effort constraint collection}

Our previous experiments indicate that span constraints are an effective way to improve unsupervised parsing. How can we leverage this method to improve unsupervised parsing in a real world setting? We explore two methods for easily finding span constraints (see Table \ref{tab:real_world}).

We find that PMI is effective as a lexicon, but not as much as the gazetteer. PMI provides more constraints than the gazetteer, but the constraints disagree more frequently with the ground truth structure and a smaller percentage of spans align exactly with the ground truth. The gazetteer approach is better than using CCKY with ground truth entity spans, despite using less than half as many constraints that only align exactly with the ground truth nearly half the time. We use gazetteer in only the most naive way via exact string matching, so we suspect that a more sophisticated yet still high precision approach (e.g. approximate string match) would have more hits and provide more benefit.

For both PMI and Gazetteer, we found that $\textsc{NCBL}$ gave the best performance.

\begin{table}[t]
\setlength\tabcolsep{6pt}
\begin{center}
\resizebox{\columnwidth}{!}{%
\begin{tabular}{l | r | r r r r r r }
\toprule
& \multicolumn{1}{c|}{WSJ} & \multicolumn{6}{c}{Constraints}  \\
           & F1   & EM & C & $n^z$ & R$^{pre}_{train}$ & R$^{post}_{train}$ & R$^{post}_{test}$    \\
\midrule
DIORA      & 56.8 & $\emptyset$ & $\emptyset$ & $\emptyset$ & $\emptyset$ & $\emptyset$ & $\emptyset$          \\ 
\midrule
+Entity    & 61.9 & 96.3 & 1.9 & 58,075 & 79.3 & 98.9 & 96.4          \\ 
\midrule
+PMI       & 57.8 & 43.9 & 7.4 & 31,965 & 75.3 & 94.4 & 90.0          \\
+Gazetteer & 58.8 & 51.3 & 5.0 & 22,354 & 80.2 & 97.0 & 93.4          \\
\bottomrule
\end{tabular}
}
\end{center}
\vspace{-0.4cm}
\caption{Parsing F1 on PTB. The max F1 across random seeds is measured on the test set. The corresponding span recall is shown on the Ontonotes $_{train}$ and $_{test}$ data before (R$^{pre}$) and after (R$^{post}$) training. The first row shows DIORA performance. Following rows show performance using distant supervision.  EM: Exact Match (percent of span constraints that are also constituents); C: Crossing (percent of span constraints that cross a constituent); $n^z$: Number of span constraints. The constraint-based metrics are not applicable to DIORA and marked with $\emptyset$.}
\label{tab:real_world}
\end{table}

\subsection{Impact on consistent convergence}

We find that using constraints with PS-SVM considerably decreases the variance on performance compared with previous baselines.\footnote{Although, most previous work does not explicitly report the standard deviation (STDEV), we can use the mean, max, and number of trials to compute the lower bound on STDEV. This yields 2.5 (Compound PCFG), 3.2 (S-DIORA), and 1.6 (RoBERTa). In contrast, our best setting has STDEV 0.6.} This is not surprising given that latent tree learning (i.e. unsupervised parsing) can converge to many equally viable parsing strategies. By using constraints, we are guiding optimization to converge to a point more aligned with the desired downstream task.

% TODO(AD): Need to update this section to reflect the new figures/captions.
% TODO(AD): Update all the figure styles.
\subsection{Are entity spans sufficient as constraints? \label{sec:synthetic}}

\begin{figure}[t]
\centering
\resizebox{\columnwidth}{!}{%
    \begin{subfigure}[b]{0.49\linewidth} \centering %
        \includegraphics[width=\linewidth]{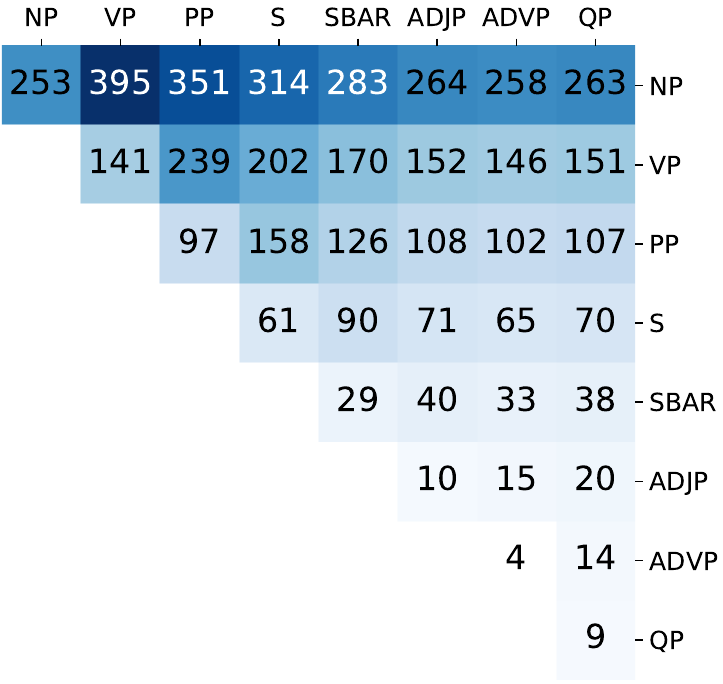}
        \caption{Span Constraint Count.}
    \end{subfigure} \hfill
    \begin{subfigure}[b]{0.49\linewidth} \centering %
        \includegraphics[width=\linewidth]{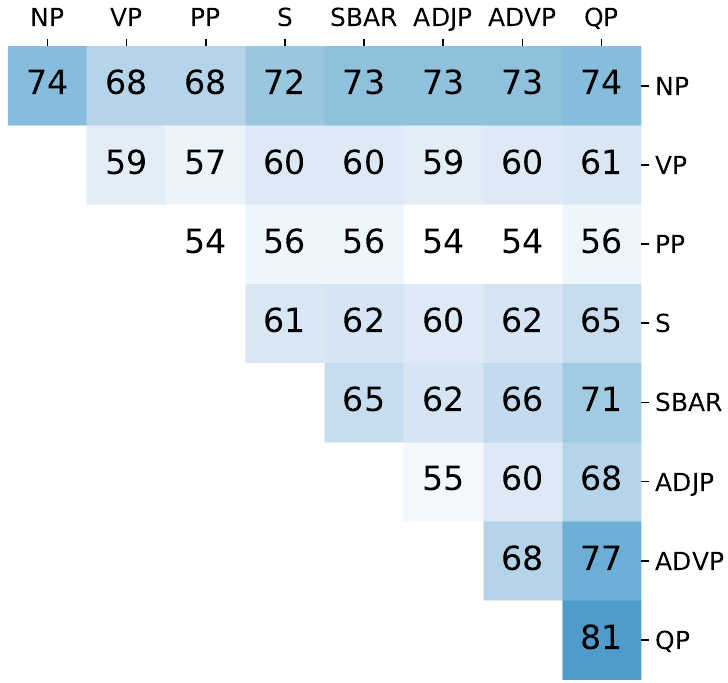}
        \caption{Initial Span Recall.}
    \end{subfigure}    
}
    \hfill
\resizebox{\columnwidth}{!}{%
    \begin{subfigure}[b]{0.49\linewidth} \centering %
        \includegraphics[width=\linewidth]{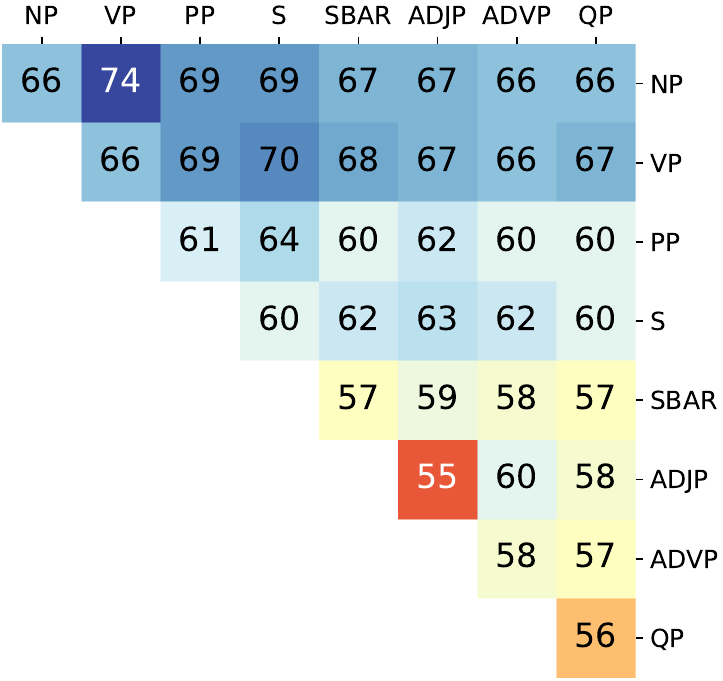}
        \caption{Parsing F1.}
    \end{subfigure} \hfill
    \begin{subfigure}[b]{0.49\linewidth} \centering %
        \includegraphics[width=\linewidth]{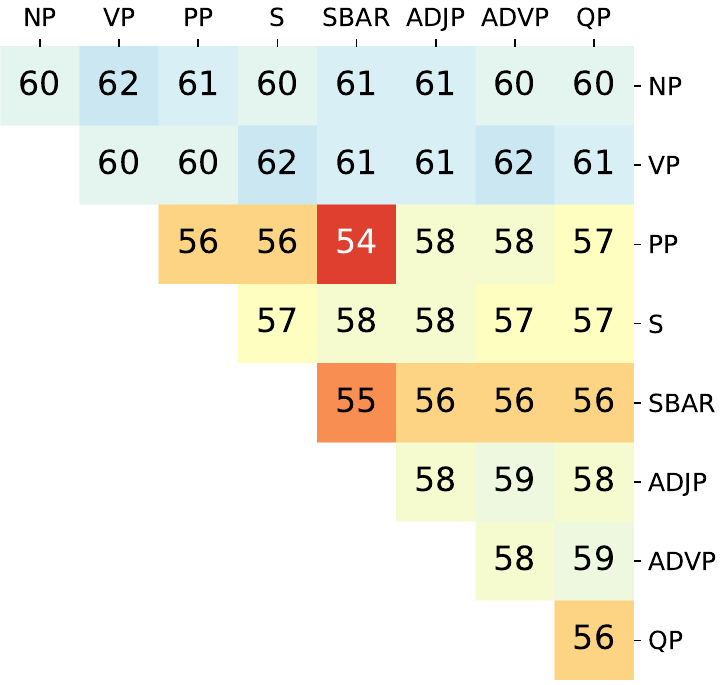}
        \caption{Parsing F1 (restricted).}
    \end{subfigure}    
}
\caption{Various statistics when using 1 or 2 constituent types as span constraints on the WSJ training and validation data. (a): The count of each span constraint in the training data (in thousands). (b): The percent of span constraints captured (i.e. span recall) in the validation data. (c): Parsing F1 on the validation data when using the span constraints with PS-SVM. (d): Parsing F1 on the validation data when using PS-SVM, although span constraints have been restricted to match the frequency and nesting behavior of entities.}
    \label{fig:onto_span_constraints}
\end{figure}

Given that DIORA already captures a large percentage of span constraints represented by entities, it is somewhat surprising that including them gives any F1 improvement. That being said, it is difficult to know \textit{a priori} which span constraints are most beneficial and how much improvement to expect. To help understand the benefits of different types of span constraints, we derived synthetic constraints using the most frequent constituent types from ground truth parse trees in Ontonotes (see Figure \ref{fig:onto_span_constraints}). The constraints extracted this way look very different from the entity constraints in that they often are nested and in general are much more frequent. To make a more fair comparison we prevent nesting and downsample to match the frequency of the entity constraints (see Figure \ref{fig:onto_span_constraints}d).

From these experiments, we can see NP or VP combined with other constraints usually lead to the best parsing performance (Figure \ref{fig:onto_span_constraints}c). This is the case even if DIORA had relatively low span recall on a different constraint type (Figure \ref{fig:onto_span_constraints}b). A reasonable hypothesis is that simply having more constraints leads to better performance, which mirrors the result that the settings with the most constraints perform better overall (Figure \ref{fig:onto_span_constraints}a). When filtered to match the shape and frequency of entity constraints, we see that performance based on NP constraints is nearly the same as with entities (Figure \ref{fig:onto_span_constraints}d). This suggests that entity spans are effective as constraints with respect to other types of constraints, but that in general we should aim to gather as many constraints as possible.

% 
% \begin{table}[ht]
% \setlength\tabcolsep{6pt}
% \begin{center}
% \resizebox{\columnwidth}{!}{%
% \begin{tabular}{l | r r r | r r}
% \toprule
% & \multicolumn{3}{c|}{WSJ} & \multicolumn{2}{c}{Constraints}  \\
%  & F1$_{\mathrm{ALL}}$ & F1$_{\leq 10}$ & F1$_{\leq 40}$ &  R$_{train}$ & R$_{test}$  \\
% \midrule
% DIORA & 56.8 & 60.8 & 57.0 & - & 81.8  \\ 
% \midrule
% +Entity & 61.9 & 64.7 & 62.1 & - & 96.4  \\ 
% % +NP+VP+PP & 69.9 & - & - & - & - &  - \\ 
% \midrule
% % +PMI & 56.5 & 68.1 & 49.5 & - & - & - \\ 
% % +PoS &  &  &  & - & - &  - \\ 
% % +Gazetteer & 58.0 & 69.9 & 50.9 & - & - & - \\
% +PMI & 56.7 & 60.5 & 56.9 & - & 88.8 \\
% +Gazetteer & 58.7 & 63.6 & 58.9 & - & 95.1 \\
% \bottomrule
% \end{tabular}
% }
% \end{center}
% \vspace{-0.4cm}
% \caption{Parsing F1 on Penn Treebank. Using the Ontonotes validation split we measure recall on constraints before (R$^{pre}$) and after (R$^{post}$) training. The first row shows DIORA performance. Following rows show performance using Partially Structured SVM (PS-SVM) with various types of constraints. The settings with PMI and Gazetteer indicate the benefit of PS-SVM in a real world setting.}
% \label{tab:real_world}
% \end{table}

\begin{table}[ht!]
\setlength\tabcolsep{4pt}
\begin{center}
\resizebox{0.7\columnwidth}{!}{%
\begin{tabular}{ l | r | r r}
\toprule
& CRAFT & \multicolumn{2}{c}{Constraints}  \\
& F1 & R$_{train}$  & R$_{test}$ \\
\midrule
\multicolumn{1}{l|}{UB}    & 85.4 & 82.8 & 79.2  \\  
\midrule
\multicolumn{1}{l|}{DIORA}        & 50.7  & 47.4 & 44.8 \\ 
\multicolumn{1}{l|}{DIORA$_{ft}$} & 55.8  & 72.4 & 65.9  \\ 
\midrule
+CCKY   & 56.2 & 99.0 & 98.6  \\ 
+PS-SVM & \textbf{56.8} & 91.1 & 85.3  \\ 
\bottomrule
\end{tabular}
}
\end{center}
\vspace{-0.4cm}
\caption{Parsing F1 and Span Recall on CRAFT. The max F1 across random seeds is measured on the test set. DIORA$_{ft}$: Fine-tuned on word prediction to assist domain transfer. UB: The upper bound on performance measured by binarizing the ground truth tree.}
\label{tab:craft}
\end{table}

\subsection{Case Study: Parsing Biomedical Text}

The most impactful domain for our method would be unsupervised parsing in a domain where full constituency tree annotation is very expensive, and span constraints are relatively easy to acquire. For this reason, we run experiments using the CRAFT corpus \cite{Verspoor2011ACO}, which contains text from biomedical research. The results are summarized in Tables \ref{tab:craft} and \ref{tab:craft_breakdown}.

\subsubsection{Domain Adaptation: Fine-tuning through Word Prediction \label{sec:craft_adapt}}

Although CRAFT and PTB are both in English, the text in biomedical research is considerably different compared with text in the newswire domain. When we evaluate the pre-trained DIORA model on the CRAFT test set, we find it achieves 50.7 F1. By simply fine-tuning the DIORA model on biomedical research text using only the word-prediction objective ($J_{rec}$) we can improve this performance to 55.8 F1 (+5.1 F1; DIORA$_{ft}$ in Table \ref{tab:craft}). This observation accentuates a beneficial property about unsupervised parsing models like DIORA: for domain adaptation, simply continue training on data from the target domain, which is possible because the word-prediction objective does not require label collection, unlike supervised models.

% This demonstrates that fine-tuning on the text from your test domain can be an effective strategy for domain adaptation for unsupervised parsing using models like DIORA. 

\subsubsection{Incorporating Span Constraints}

We use the ground truth entity annotation in the CRAFT training data as a source of distant supervision and continue training DIORA using the PS-SVM objective. By incorporating span constraints this way, we see that parsing performance on the test set improves from $55.8 \rightarrow 56.8$ (+1 F1).

For CRAFT, we used grid search over a small set of hyperparameters including loss variants and found that $\textsc{Structured Ramp}$ performed best.

\newpage

\paragraph{Performance by Sentence Type} In Table \ref{tab:craft_breakdown} we report parsing results bucketed by sentence-type determined by the top-most constituent label. In general, across almost all sentence types, simply constraining the DIORA output to incorporate known spans boosts F1 performance. Training with the PS-SVM objective usually improves F1 further, although the amount depends on the sentence type.

% For example, we see the largest relative improvement in TITLE and FRAG --- we also see improvements in SBARQ and SQ even though all the constraints were already satisfied prior to training with PS-SVM. Sentences with NP as the top-most constituent see an improvement when using PS-SVM compared with DIORA, but using the hard constraints (+CCKY) actually does better. For these NP sentences, we notice that the recall on span constraints improves from DIORA, but at 52.9 it is still very low compared with other sentence types.

% TODO(AD): The second issue is plausible, but really could be made more concrete or even described in an entirely different way. Essentially, the training setup we use for newswire did not transfer well. What we need is the `bag of tricks' often used with BERT fine-tuning that leads to good performance.

% \textcolor{red}{(in progress)}

\paragraph{Challenging NP-type Sentences} We observe especially low span-recall for sentences with NP as the top-most constituent (Table \ref{tab:craft_breakdown}). These are short sentences that exhibit  domain-specific structure. Here is a typical sentence and ground truth parse for that case:

% \begin{lstlisting}
% (
%   ((HIF - 1$\alpha$) KO)
%   -
%   (
%     (skeletal - muscle)
%     (HIF - 1$\alpha$)
%     knockout mouse))
% \end{lstlisting}

\begin{quote} 
    \textit{((HIF - 1$\alpha$) KO) - ((skeletal - muscle) \\ \hspace*{\parindent} (HIF - 1$\alpha$) knockout mouse)}
\end{quote}

\noindent
Various properties of the above sentence make it difficult to parse. For instance, the sentence construction lacks syntactic cues and there is no verb in the sentence. There is also substantial ambiguity with respect to hyphenation, and the second hyphen is acting as a colon. These properties make it difficult to capture the spans \textit{(skeletal - muscle)} or the second \textit{(HIF - 1$\alpha$)} despite being constraints.

\begin{table}[t!]
\setlength\tabcolsep{4pt}
\begin{center}
\resizebox{\columnwidth}{!}{%
\begin{tabular}{l | r r | r r | r r | r r}
\toprule
& & & \multicolumn{2}{c|}{DIORA} & \multicolumn{2}{c|}{+CCKY} & \multicolumn{2}{c}{+PS-SVM}  \\
& $n$ & $n^z$ & F1 & R$^z$ & F1 & R$^z$ & F1 & R$^z$  \\
\midrule
CAPTION & 1857 & 1579 & 55.7 & 67.6 & 56.0 & 98.3 & 56.0 & 86.1\\
HEADING & 1149 & 201 & 72.0 & 59.2 & 72.8 & 96.5 & 73.5 & 83.1\\
TITLE & 29 & 31 & 51.4 & 58.1 & 53.8 & 96.8 & 55.4 & 71.0\\
CIT & 3 & 0 & 40.0 & $\emptyset$ & 40.0 & $\emptyset$ & 40.0 & $\emptyset$\\
S-IMP & 1 & 0 & 36.8 & $\emptyset$ & 36.8 & $\emptyset$ & 31.6 & $\emptyset$\\
\midrule
S & 5872 & 5140 & 53.9 & 65.9 & 54.2 & 98.8 & 54.9 & 85.4\\
NP & 136 & 34 & 37.1 & 41.2 & 40.6 & 100.0 & 44.1 & 52.9\\
FRAG & 39 & 52 & 49.3 & 71.2 & 49.0 & 100.0 & 51.8 & 84.6\\
SINV & 6 & 7 & 50.7 & 42.9 & 47.9 & 85.7 & 46.3 & 57.1\\
SBARQ & 5 & 1 & 49.5 & 100.0 & 49.5 & 100.0 & 55.1 & 100.0\\
SQ & 2 & 1 & 28.0 & 100.0 & 28.0 & 100.0 & 32.9 & 100.0\\
\bottomrule
\end{tabular}
}
\end{center}
\vspace{-0.4cm}
\caption{Parsing F1 on CRAFT test set from the best model bucketed by the sentence's top-most constituent type. $n$: Count of sentences. $n^z$: Count of constraints. R$^z$: Recall on constraints. $\emptyset$: Indicates no constraints.}
\label{tab:craft_breakdown}
\end{table}

\subsubsection{Parsing of PTB vs. CRAFT }

As mentioned in \S\ref{sec:craft_adapt}, there is considerable difference in the text between PTB and CRAFT. It follows that there would be a difference in difficulty when parsing these two types of data. After running the parser from \citet{kitaev-klein-2018-constituency} on each dataset, it appears CRAFT is more difficult to parse than PTB. For CRAFT, the unlabeled parsing F1 is 81.3 and the span recall for entities is 37.6. For PTB, the unlabeled parsing F1 is 95.

\section{Related Work }

% \paragraph{Latent tree learning} Various recent works induce structures that can be used as constituency trees \cite{shen2018ordered,kim2019compound}. Although typically evaluated against PTB, one can not expect ground truth parse trees to be fully recovered for the same reason clustering algorithms do not always recover image classes. In contrast, with distant supervision we can guide convergence of unsupervised learning to a desirable point.

\paragraph{Learning from Partially Labeled Corpora} \citet{pereira-schabes-1992-inside} modify the inside-outside algorithm to respect span constraints. Similar methods have been explored for training CRFs \cite{culotta-mccallum-2004-confidence,Bellare2007LearningEF}. Rather than modify the weight assignment in DIORA, which is inspired by the inside-outside algorithm, we supervise the tree predicted from the inside-pass.

Concurrent work to ours in distant supervision trains RoBERTa for constituency parsing using answer spans from question-answering datasets and wikipedia hyperlinks \cite{shi2021learning}. Although effective, their approach depends entirely on the set of constraints. In contrast, PS-SVM enhances DIORA, which is a model that outputs a parse tree without any supervision.

The span constraints in this work are derived from external resources, and do not necessarily match the parse tree. Constraints may conflict with the parse, which is why CCKY can be less than 100 span recall in Table \ref{tab:craft}. This approach to model training is often called ``distant supervision'' \cite{mintz-etal-2009-distant,shi2021learning}. In contrast, ``partial supervision'' implies gold partial labels are available, which we explore as synthetic data (\S\ref{sec:synthetic}), but in general do not make this assumption.

% PR - modifies the probability 
% CoDL - similar to our method but don't think it has been tested on totally unsupervised setup. 
% A common infrastructure across tasks  made it easy to train multiple task objectives simultaneously with a shared parameters. Perhaps the first attempt of this kind is by Collobert et al. (2011)

% \jy{[Marked phrases in blue that I am ambiguious to include]}

\paragraph{Joint Supervision} 

% refernce sentence for multi-task learning definition:
% Multitask Learning is an approach to inductive transfer that improves generalization by using the domain information contained in the training signals of related tasks as an inductive bias. It does this by learning tasks in parallel while using a shared representation; what is learned for each task can help other tasks be learned better.
% Perhaps, the most naive way to inject additional information of span constraint 
% While we take explicit use of constraint in this paper, perhaps more implicit way of incorporating constraints would be through multi-task learning (Caruana cite). 
% While this approach does not capture explicit relation across the tasks,it showed successful improvement in performance \jy{ through inductive transfer } in neural NLP from as early as (Collobert cite) to (Swabha cite). 
An implicit way to incorporate constraints is through multi-task learning (MTL; \citealp{caruana1997multitask}).
Even when relations between the tasks are not modeled explicitly, MTL has shown promise throughout a range of text processing tasks with neural models \cite{Collobert2008AUA,Swayamdipta2018SyntacticSF,Kuncoro2020SyntacticSD}.
Preliminary experiments with joint NER did not improving parsing results. This is in-line with DIORA’s relative weakness in representing fine-grained entity types. Modifications of DIORA to improve its semantic representation may prove to make joint NER more viable.
% \jy{Especially Swabha et al showed that by simply learning a main task, which were span-based tasks such as SRL or Coreference resolution, together with subsidiary task, parsing, on a shared parameter results in improvement of the main task. }

% {\color{red} However, in our preliminary experiment of joint learning of NER loss and DIORA's reconstruction loss did not lead to positive gains in perfornace; in contrast, it degraded the performance of parsing a bit while the NER span constraints are quite relevant for parsing. }

\paragraph{Constraint Injection Methods} 

There exists a rich literature in constraint injection \cite{Ganchev2010PosteriorRF,chang2012structured} . Both methods are based on Expectation Maximization (EM) algorithm \cite{Dempster1977MaximumLF} where the constraint is injected in the E-step of calculating the posterior distribution \cite{samdani2012unified}. Another line of work focuses injecting constraint in the M-step \cite{GradientBased, mehta2018towards} by reflecting the degree of constraint satisfaction of prediction as the weight of the gradient.  Our approach is similar to \citet{chang2012structured} as we select the highest scoring output that satisfies constraints and learn from it. $\text{PS-SVM}_{\textsc{Rescale}}$ is based on \citet{GradientBased}.
% We also examine whether \citet{GradientBased}'s approach is beneficial on our PS-SVM formulation by experimenting with the ${\textsc{PS-SVM}_\textsc{Rescale}}$.
% \footnote{Chang et al. takes cross entropy loss using the $y^+$ as there are distribution over labels. In contrast, our approach does not have labels in between and only have access to aggregate tree score and we use SSVM loss to reduce the margin when the score of negative tree $y^-$ is larger than $y^+$. \jy{This can be removed, wrote it in my thinking process}}

The aforementioned constraint injection methods were usually used as an added loss to the supervised loss function. In this work, we show that the distant supervision through constraint injection is beneficial for unsupervised setting as well. 

\paragraph{Structural SVM with Latent Variables} The PS-SVM loss we introduce in this work can be loosely thought of as an application-specific instantiation of Structural SVM with Latent Variables \cite{Yu2009LearningSS}. Various works have extended Structural SVM with Latent Variables to incorporate constraints for tasks such as sequence labeling \cite{Yu2012Prior} and co-reference resolution \cite{chang-etal-2013-constrained}, although none we have seen focus on unsupervised constituency parsing. Perhaps a more clear distinction is that \citet{Yu2009LearningSS} focuses on latent variables within supervised tasks, and PS-SVM is meant to improve convergence of an unsupervised learning algorithm (i.e., DIORA).

\paragraph{Additional Related Work} In Appendix \ref{sec:app_related_work} we list additional work in unsupervised parsing not already mentioned.
% , including works that use alternative constraints such as limited recursion depth.
\section{Conclusion}

In this work, we present a method for enhancing DIORA with distant supervision from span constraints. We call this approach Partially Structured SVM (PS-SVM). We find that span constraints based on entities are effective at improving parsing performance of DIORA on English newswire data (+5.1 F1 using ground truth entities, or +2 F1 using a gazetteer). Furthermore, we show PS-SVM is also effective in the domain of biomedical text (+1 F1 using ground truth entities). Our detailed analysis shows that entities are effective as span constraints, giving equivalent benefit as a similar amount of NP-based constraints. We hope our findings will help ``bridge the gap'' between supervised and unsupervised parsing. 
% In future work, intend to use data augmentation to expand the constraint set and leverage stronger models (i.e. S-DIORA; \citealp{drozdov2020sdiora}) to further improve performance.

\section*{Broader Impact}

We hope our work will increase the availability of parse tree annotation for low-resource domains, generated in an unsupervised manner. Compared with full parse tree annotation, span constraints can be acquired at reduced cost or even automatically extracted.

The gazetteer used in our experiments is automatically extracted from Wikipedia, and our experiments are only for English, which is the language with by far the most Wikipedia entries. Although, similarly sized gazetteers may be difficult to attain in other languages, \citet{mikheev-etal-1999-named} point out larger gazetteers do not necessarily boost performance, and gazetteers have already proven effective in low-resource domains \cite{rijhwani-etal-2020-soft}. In any case, we use gazetteers in the most naive way by finding exact text matches. When extending our approach to other languages, an entity recognition model may be a suitable replacement for the gazetteer.

\section*{Acknowledgements}

We are grateful to our colleagues at UMass NLP and the anonymous reviewers for feedback on drafts of this work. This work was supported in part by the Center for Intelligent Information Retrieval, in part by the Chan Zuckerberg Initiative, in part by the IBM Research AI through the AI Horizons Network, and in part by the National Science Foundation (NSF) grant numbers DMR-1534431, IIS-1514053, CNS-0958392, and IIS-1955567. Any opinions, findings and conclusions or recommendations expressed in this material are those of the authors and do not necessarily reflect those of the sponsor.

\bibliography{bibliography}

\begin{thebibliography}{69}
\expandafter\ifx\csname natexlab\endcsname\relax\def\natexlab#1{#1}\fi

\bibitem[{Akoury et~al.(2019)Akoury, Krishna, and
  Iyyer}]{akoury-etal-2019-syntactically}
Nader Akoury, Kalpesh Krishna, and Mohit Iyyer. 2019.
\newblock \href {https://doi.org/10.18653/v1/P19-1122} {Syntactically
  supervised transformers for faster neural machine translation}.
\newblock In \emph{Proceedings of the 57th Annual Meeting of the Association
  for Computational Linguistics}, pages 1269--1281, Florence, Italy.
  Association for Computational Linguistics.

\bibitem[{Bellare and McCallum(2007)}]{Bellare2007LearningEF}
Kedar Bellare and Andrew McCallum. 2007.
\newblock Learning extractors from unlabeled text using relevant databases.
\newblock In \emph{Proceedings of the 2007 {AAAI} Workshop on information
  integration on the web}.

\bibitem[{Bies et~al.(1995)Bies, Ferguson, Katz, MacIntyre, Tredinnick, Kim,
  Marcinkiewicz, and Schasberger}]{bies1995bracketing}
Ann Bies, Mark Ferguson, Karen Katz, Robert MacIntyre, Victoria Tredinnick,
  Grace Kim, Mary~Ann Marcinkiewicz, and Britta Schasberger. 1995.
\newblock Bracketing guidelines for {Treebank} {II} style {Penn Treebank}
  project.
\newblock Technical report, Department of Linguistics, University of
  Pennsylvania.

\bibitem[{Brill et~al.(1990)Brill, Magerman, Marcus, and
  Santorini}]{brill-etal-1990-deducing}
Eric Brill, David Magerman, Mitchell Marcus, and Beatrice Santorini. 1990.
\newblock \href {https://www.aclweb.org/anthology/H90-1055} {Deducing
  linguistic structure from the statistics of large corpora}.
\newblock In \emph{Speech and Natural Language: Proceedings of a Workshop Held
  at Hidden Valley, {P}ennsylvania, June 24-27,1990}.

\bibitem[{Cao et~al.(2020)Cao, Kitaev, and Klein}]{cao-etal-2020-unsupervised}
Steven Cao, Nikita Kitaev, and Dan Klein. 2020.
\newblock \href {https://doi.org/10.18653/v1/2020.emnlp-main.389} {Unsupervised
  parsing via constituency tests}.
\newblock In \emph{Proceedings of the 2020 Conference on Empirical Methods in
  Natural Language Processing (EMNLP)}, pages 4798--4808, Online. Association
  for Computational Linguistics.

\bibitem[{Carroll and Charniak(1992)}]{Carroll1992TwoEO}
Glenn Carroll and Eugene Charniak. 1992.
\newblock Two experiments on learning probabilistic dependency grammars from
  corpora.
\newblock Technical report, Dept. of Computer Science, Brown University.

\bibitem[{Caruana(1997)}]{caruana1997multitask}
Rich Caruana. 1997.
\newblock Multitask learning.
\newblock \emph{Machine learning}, 28(1):41--75.

\bibitem[{Chang et~al.(2013)Chang, Samdani, and
  Roth}]{chang-etal-2013-constrained}
Kai-Wei Chang, Rajhans Samdani, and Dan Roth. 2013.
\newblock \href {https://aclanthology.org/D13-1057} {A constrained latent
  variable model for coreference resolution}.
\newblock In \emph{Proceedings of the 2013 Conference on Empirical Methods in
  Natural Language Processing}, pages 601--612, Seattle, Washington, USA.
  Association for Computational Linguistics.

\bibitem[{Chang et~al.(2012)Chang, Ratinov, and Roth}]{chang2012structured}
Ming-Wei Chang, Lev Ratinov, and Dan Roth. 2012.
\newblock Structured learning with constrained conditional models.
\newblock \emph{Machine learning}, 88(3):399--431.

\bibitem[{Chapelle et~al.(2009)Chapelle, B., Teo, Le, and
  Smola}]{do_structured_ramp}
Olivier Chapelle, Chuong B., Choon Teo, Quoc Le, and Alex Smola. 2009.
\newblock \href
  {https://proceedings.neurips.cc/paper/2008/file/6bc24fc1ab650b25b4114e93a98f1eba-Paper.pdf}
  {Tighter bounds for structured estimation}.
\newblock In \emph{Advances in Neural Information Processing Systems},
  volume~21. Curran Associates, Inc.

\bibitem[{Clark(2001)}]{Clark2001UnsupervisedIO}
Alexander Clark. 2001.
\newblock Unsupervised induction of stochastic context-free grammars using
  distributional clustering.
\newblock In \emph{CoNLL}.

\bibitem[{Cohen et~al.(2017)Cohen, Verspoor, Fort, Funk, Bada, Palmer, and
  Hunter}]{Cohen2017TheCR}
Kevin~Bretonnel Cohen, Karin Verspoor, Kar{\"e}n Fort, Christopher Funk,
  Michael Bada, Martha Palmer, and Lawrence Hunter. 2017.
\newblock The colorado richly annotated full text {(CRAFT)} corpus: Multi-model
  annotation in the biomedical domain.
\newblock In \emph{Handbook of Linguistic Annotation}, page 1379 – 1394.
  Springer.

\bibitem[{Collobert and Weston(2008)}]{Collobert2008AUA}
Ronan Collobert and J.~Weston. 2008.
\newblock A unified architecture for natural language processing: deep neural
  networks with multitask learning.
\newblock In \emph{ICML '08}.

\bibitem[{Culotta and McCallum(2004)}]{culotta-mccallum-2004-confidence}
Aron Culotta and Andrew McCallum. 2004.
\newblock \href {https://www.aclweb.org/anthology/N04-4028} {Confidence
  estimation for information extraction}.
\newblock In \emph{Proceedings of {HLT}-{NAACL} 2004: Short Papers}, pages
  109--112, Boston, Massachusetts, USA. Association for Computational
  Linguistics.

\bibitem[{Dempster et~al.(1977)Dempster, Laird, and
  Rubin}]{Dempster1977MaximumLF}
A.~Dempster, N.~Laird, and D.~Rubin. 1977.
\newblock Maximum likelihood from incomplete data via the {EM} - algorithm plus
  discussions on the paper.
\newblock \emph{Journal of the Royal Statistical Society: Series B
  (Methodological)}.

\bibitem[{Drozdov et~al.(2020)Drozdov, Rongali, Chen, O{'}Gorman, Iyyer, and
  McCallum}]{drozdov2020sdiora}
Andrew Drozdov, Subendhu Rongali, Yi-Pei Chen, Tim O{'}Gorman, Mohit Iyyer, and
  Andrew McCallum. 2020.
\newblock Unsupervised parsing with {S-DIORA}: Single tree encoding for deep
  inside-outside recursive autoencoders.
\newblock In \emph{Empirical Methods in Natural Language Processing (EMNLP)}.

\bibitem[{Drozdov et~al.(2019)Drozdov, Verga, Yadav, Iyyer, and
  McCallum}]{Diora}
Andrew Drozdov, Patrick Verga, Mohit Yadav, Mohit Iyyer, and Andrew McCallum.
  2019.
\newblock Unsupervised latent tree induction with deep inside-outside recursive
  autoencoders.
\newblock In \emph{NAACL-HLT}.

\bibitem[{Finkel and Manning(2009)}]{Finkel2009JointPA}
Jenny~Rose Finkel and Christopher~D. Manning. 2009.
\newblock \href {https://www.aclweb.org/anthology/N09-1037/} {Joint parsing and
  named entity recognition}.
\newblock In \emph{Human Language Technologies: Conference of the North
  American Chapter of the Association of Computational Linguistics,
  Proceedings, May 31 - June 5, 2009, Boulder, Colorado, {USA}}, pages
  326--334. The Association for Computational Linguistics.

\bibitem[{Ganchev et~al.(2010)Ganchev, Graça, Gillenwater, and
  Taskar}]{Ganchev2010PosteriorRF}
Kuzman Ganchev, Jo{\~a}o Graça, Jennifer Gillenwater, and B.~Taskar. 2010.
\newblock Posterior regularization for structured latent variable models.
\newblock \emph{J. Mach. Learn. Res.}, 11:2001--2049.

\bibitem[{Gimpel and Smith(2012)}]{gimpel-smith-2012-structured}
Kevin Gimpel and Noah~A. Smith. 2012.
\newblock \href {https://www.aclweb.org/anthology/N12-1023} {Structured ramp
  loss minimization for machine translation}.
\newblock In \emph{Proceedings of the 2012 Conference of the North {A}merican
  Chapter of the Association for Computational Linguistics: Human Language
  Technologies}, pages 221--231, Montr{\'e}al, Canada. Association for
  Computational Linguistics.

\bibitem[{Haghighi and Klein(2006)}]{haghighi-klein-2006-prototype}
Aria Haghighi and Dan Klein. 2006.
\newblock Prototype-driven grammar induction.
\newblock In \emph{Proceedings of the 21st International Conference on
  Computational Linguistics and 44th Annual Meeting of the Association for
  Computational Linguistics}.

\bibitem[{Htut et~al.(2018)Htut, Cho, and Bowman}]{htut-etal-2018-grammar}
Phu~Mon Htut, Kyunghyun Cho, and Samuel Bowman. 2018.
\newblock \href {https://doi.org/10.18653/v1/W18-5452} {Grammar induction with
  neural language models: An unusual replication}.
\newblock In \emph{Proceedings of the 2018 {EMNLP} Workshop {B}lackbox{NLP}:
  Analyzing and Interpreting Neural Networks for {NLP}}, pages 371--373,
  Brussels, Belgium. Association for Computational Linguistics.

\bibitem[{Iyyer et~al.(2018)Iyyer, Wieting, Gimpel, and
  Zettlemoyer}]{iyyer-etal-2018-adversarial}
Mohit Iyyer, John Wieting, Kevin Gimpel, and Luke Zettlemoyer. 2018.
\newblock \href {https://doi.org/10.18653/v1/N18-1170} {Adversarial example
  generation with syntactically controlled paraphrase networks}.
\newblock In \emph{Proceedings of the 2018 Conference of the North {A}merican
  Chapter of the Association for Computational Linguistics: Human Language
  Technologies, Volume 1 (Long Papers)}, pages 1875--1885, New Orleans,
  Louisiana. Association for Computational Linguistics.

\bibitem[{Jin et~al.(2018)Jin, Doshi-Velez, Miller, Schuler, and
  Schwartz}]{jin_depth_pcfg}
Lifeng Jin, Finale Doshi-Velez, Timothy Miller, William Schuler, and Lane
  Schwartz. 2018.
\newblock \href {https://doi.org/10.1162/tacl_a_00016} {{Unsupervised Grammar
  Induction with Depth-bounded PCFG}}.
\newblock \emph{Transactions of the Association for Computational Linguistics},
  6:211--224.

\bibitem[{Kasami(1965)}]{CKY1}
T.~Kasami. 1965.
\newblock An efficient recognition and syntax analysis algorithm for
  context-free languages.
\newblock Technical Report AFCRL-65-758, Air Force Cambridge Research
  Laboratory, Bedford, MA$\dag$.

\bibitem[{Kim et~al.(2019{\natexlab{a}})Kim, Dyer, and Rush}]{kim2019compound}
Yoon Kim, Chris Dyer, and Alexander~M Rush. 2019{\natexlab{a}}.
\newblock Compound probabilistic context-free grammars for grammar induction.
\newblock In \emph{ACL}.

\bibitem[{Kim et~al.(2019{\natexlab{b}})Kim, Rush, Yu, Kuncoro, Dyer, and
  Melis}]{kim-etal-2019-unsupervised}
Yoon Kim, Alexander Rush, Lei Yu, Adhiguna Kuncoro, Chris Dyer, and G{\'a}bor
  Melis. 2019{\natexlab{b}}.
\newblock \href {https://doi.org/10.18653/v1/N19-1114} {Unsupervised recurrent
  neural network grammars}.
\newblock In \emph{Proceedings of the 2019 Conference of the North {A}merican
  Chapter of the Association for Computational Linguistics: Human Language
  Technologies, Volume 1 (Long and Short Papers)}, pages 1105--1117,
  Minneapolis, Minnesota. Association for Computational Linguistics.

\bibitem[{Kitaev and Klein(2018)}]{kitaev-klein-2018-constituency}
Nikita Kitaev and Dan Klein. 2018.
\newblock Constituency parsing with a self-attentive encoder.
\newblock In \emph{Association for Computational Linguistic (ACL)}.

\bibitem[{Klein and Manning(2004)}]{klein-manning-2004-corpus}
Dan Klein and Christopher Manning. 2004.
\newblock \href {https://doi.org/10.3115/1218955.1219016} {Corpus-based
  induction of syntactic structure: Models of dependency and constituency}.
\newblock In \emph{Proceedings of the 42nd Annual Meeting of the Association
  for Computational Linguistics ({ACL}-04)}, pages 478--485, Barcelona, Spain.

\bibitem[{Klein and Manning(2001)}]{Klein2001NaturalLG}
Dan Klein and Christopher~D. Manning. 2001.
\newblock Natural language grammar induction using a constituent-context model.
\newblock In \emph{NeurIPS}.

\bibitem[{Kuncoro et~al.(2020)Kuncoro, Kong, Fried, Yogatama, Rimell, Dyer, and
  Blunsom}]{Kuncoro2020SyntacticSD}
Adhiguna Kuncoro, Lingpeng Kong, Daniel Fried, Dani Yogatama, Laura Rimell,
  Chris Dyer, and Phil Blunsom. 2020.
\newblock Syntactic structure distillation pretraining for bidirectional
  encoders.
\newblock \emph{Transactions of the Association for Computational Linguistics},
  8:776--794.

\bibitem[{Lari and Young(1990)}]{lari1990estimation}
Karim Lari and Steve~J Young. 1990.
\newblock The estimation of stochastic context-free grammars using the
  inside-outside algorithm.
\newblock \emph{Computer speech \& language}, 4(1):35--56.

\bibitem[{Lee et~al.(2019)Lee, Mehta, Wick, Tristan, and
  Carbonell}]{GradientBased}
Jay~Yoon Lee, Sanket~Vaibhav Mehta, Michael~L. Wick, Jean{-}Baptiste Tristan,
  and Jaime~G. Carbonell. 2019.
\newblock \href {https://doi.org/10.1609/aaai.v33i01.33014147} {Gradient-based
  inference for networks with output constraints}.
\newblock In \emph{The Thirty-Third {AAAI} Conference on Artificial
  Intelligence, {AAAI} 2019, The Thirty-First Innovative Applications of
  Artificial Intelligence Conference, {IAAI} 2019, The Ninth {AAAI} Symposium
  on Educational Advances in Artificial Intelligence, {EAAI} 2019, Honolulu,
  Hawaii, USA, January 27 - February 1, 2019}, pages 4147--4154. {AAAI} Press.

\bibitem[{Liu et~al.(2019{\natexlab{a}})Liu, Yao, and
  Lin}]{liu-etal-2019-towards}
Tianyu Liu, Jin-Ge Yao, and Chin-Yew Lin. 2019{\natexlab{a}}.
\newblock \href {https://doi.org/10.18653/v1/P19-1524} {Towards improving
  neural named entity recognition with gazetteers}.
\newblock In \emph{Proceedings of the 57th Annual Meeting of the Association
  for Computational Linguistics}, pages 5301--5307, Florence, Italy.
  Association for Computational Linguistics.

\bibitem[{Liu et~al.(2019{\natexlab{b}})Liu, Ott, Goyal, Du, Joshi, Chen, Levy,
  Lewis, Zettlemoyer, and Stoyanov}]{Liu2019RoBERTaAR}
Yinhan Liu, Myle Ott, Naman Goyal, Jingfei Du, Mandar Joshi, Danqi Chen, Omer
  Levy, Michael Lewis, Luke Zettlemoyer, and Veselin Stoyanov.
  2019{\natexlab{b}}.
\newblock Roberta: A robustly optimized bert pretraining approach.
\newblock \emph{ArXiv}, abs/1907.11692.

\bibitem[{Marcus et~al.(1993)Marcus, Santorini, and
  Marcinkiewicz}]{marcus1993building}
Mitchell Marcus, Beatrice Santorini, and Mary~Ann Marcinkiewicz. 1993.
\newblock Building a large annotated corpus of english: The penn treebank.
\newblock \emph{Computational linguistics}, 19(2):313--330.

\bibitem[{Mayhew et~al.(2019)Mayhew, Chaturvedi, Tsai, and
  Roth}]{mayhew-etal-2019-named}
Stephen Mayhew, Snigdha Chaturvedi, Chen-Tse Tsai, and Dan Roth. 2019.
\newblock \href {https://doi.org/10.18653/v1/K19-1060} {Named entity
  recognition with partially annotated training data}.
\newblock In \emph{Proceedings of the 23rd Conference on Computational Natural
  Language Learning (CoNLL)}, pages 645--655, Hong Kong, China. Association for
  Computational Linguistics.

\bibitem[{Mehta et~al.(2018)Mehta, Lee, and Carbonell}]{mehta2018towards}
Sanket~Vaibhav Mehta, Jay~Yoon Lee, and Jaime Carbonell. 2018.
\newblock \href {https://doi.org/10.18653/v1/D18-1538} {Towards semi-supervised
  learning for deep semantic role labeling}.
\newblock In \emph{Proceedings of the 2018 Conference on Empirical Methods in
  Natural Language Processing}, pages 4958--4963, Brussels, Belgium.
  Association for Computational Linguistics.

\bibitem[{Mikheev et~al.(1999)Mikheev, Moens, and
  Grover}]{mikheev-etal-1999-named}
Andrei Mikheev, Marc Moens, and Claire Grover. 1999.
\newblock \href {https://aclanthology.org/E99-1001} {Named entity recognition
  without gazetteers}.
\newblock In \emph{Ninth Conference of the {E}uropean Chapter of the
  Association for Computational Linguistics}, pages 1--8, Bergen, Norway.
  Association for Computational Linguistics.

\bibitem[{Mikolov et~al.(2013)Mikolov, Sutskever, Chen, Corrado, and
  Dean}]{Mikolov2013DistributedRO}
Tomas Mikolov, Ilya Sutskever, Kai Chen, Greg Corrado, and Jeff Dean. 2013.
\newblock Distributed representations of words and phrases and their
  compositionality.
\newblock In \emph{NeurIPS}.

\bibitem[{Mintz et~al.(2009)Mintz, Bills, Snow, and
  Jurafsky}]{mintz-etal-2009-distant}
Mike Mintz, Steven Bills, Rion Snow, and Daniel Jurafsky. 2009.
\newblock \href {https://aclanthology.org/P09-1113} {Distant supervision for
  relation extraction without labeled data}.
\newblock In \emph{Proceedings of the Joint Conference of the 47th Annual
  Meeting of the {ACL} and the 4th International Joint Conference on Natural
  Language Processing of the {AFNLP}}, pages 1003--1011, Suntec, Singapore.
  Association for Computational Linguistics.

\bibitem[{Mohan and Li(2019)}]{Mohan2019MedMentionsAL}
Sunil Mohan and Donghui Li. 2019.
\newblock Medmentions: A large biomedical corpus annotated with {UMLS}
  concepts.
\newblock In \emph{Automated Knowledge Base Construction (AKBC)}.

\bibitem[{Moore et~al.(1995)Moore, Appelt, Dowding, Gawron, and
  Moran}]{Moore1995CombiningLA}
Robert Moore, Douglas Appelt, John Dowding, J.~Mark Gawron, and Douglas Moran.
  1995.
\newblock Combining linguistic and statistical knowledge sources in
  natural-language processing for atis.
\newblock In \emph{Proceedings of the January 1995 ARPA Spoken Language Systems
  Technology Workshop}.

\bibitem[{Naradowsky(2014)}]{narad:phdthesis}
Jason Naradowsky. 2014.
\newblock \emph{Learning with Joint Inference and Latent Linguistic Structure
  in Graphical Models}.
\newblock Ph.D. thesis, University of Massachusetts Amherst.

\bibitem[{Niculae and Martins(2020)}]{pmlr-v119-niculae20a}
Vlad Niculae and Andre Martins. 2020.
\newblock \href {http://proceedings.mlr.press/v119/niculae20a.html}
  {{LP}-{S}parse{MAP}: Differentiable relaxed optimization for sparse
  structured prediction}.
\newblock In \emph{Proceedings of the 37th International Conference on Machine
  Learning}, volume 119 of \emph{Proceedings of Machine Learning Research},
  pages 7348--7359. PMLR.

\bibitem[{Pereira and Schabes(1992)}]{pereira-schabes-1992-inside}
Fernando Pereira and Yves Schabes. 1992.
\newblock \href {https://doi.org/10.3115/981967.981984} {Inside-outside
  reestimation from partially bracketed corpora}.
\newblock In \emph{30th Annual Meeting of the Association for Computational
  Linguistics}, pages 128--135, Newark, Delaware, USA. Association for
  Computational Linguistics.

\bibitem[{Phang et~al.(2018)Phang, F{\'e}vry, and Bowman}]{Phang2018SentenceEO}
Jason Phang, Thibault F{\'e}vry, and Samuel~R. Bowman. 2018.
\newblock Sentence encoders on stilts: Supplementary training on intermediate
  labeled-data tasks.
\newblock \emph{ArXiv}, abs/1811.01088.

\bibitem[{Ponvert et~al.(2011)Ponvert, Baldridge, and
  Erk}]{ponvert-etal-2011-simple}
Elias Ponvert, Jason Baldridge, and Katrin Erk. 2011.
\newblock \href {https://www.aclweb.org/anthology/P11-1108} {Simple
  unsupervised grammar induction from raw text with cascaded finite state
  models}.
\newblock In \emph{Proceedings of the 49th Annual Meeting of the Association
  for Computational Linguistics: Human Language Technologies}, pages
  1077--1086, Portland, Oregon, USA. Association for Computational Linguistics.

\bibitem[{Pradhan et~al.(2012)Pradhan, Moschitti, Xue, Uryupina, and
  Zhang}]{Conll2012}
Sameer Pradhan, Alessandro Moschitti, Nianwen Xue, Olga Uryupina, and Yuchen
  Zhang. 2012.
\newblock \href {https://www.aclweb.org/anthology/W12-4501/} {Conll-2012 shared
  task: Modeling multilingual unrestricted coreference in ontonotes}.
\newblock In \emph{Joint Conference on Empirical Methods in Natural Language
  Processing and Computational Natural Language Learning - Proceedings of the
  Shared Task: Modeling Multilingual Unrestricted Coreference in OntoNotes,
  EMNLP-CoNLL 2012, July 13, 2012, Jeju Island, Korea}, pages 1--40. {ACL}.

\bibitem[{Ratinov and Roth(2009)}]{ratinov-roth-2009-design}
Lev Ratinov and Dan Roth. 2009.
\newblock \href {https://aclanthology.org/W09-1119} {Design challenges and
  misconceptions in named entity recognition}.
\newblock In \emph{Proceedings of the Thirteenth Conference on Computational
  Natural Language Learning ({C}o{NLL}-2009)}, pages 147--155, Boulder,
  Colorado. Association for Computational Linguistics.

\bibitem[{Rijhwani et~al.(2020)Rijhwani, Zhou, Neubig, and
  Carbonell}]{rijhwani-etal-2020-soft}
Shruti Rijhwani, Shuyan Zhou, Graham Neubig, and Jaime Carbonell. 2020.
\newblock \href {https://doi.org/10.18653/v1/2020.acl-main.722} {Soft
  gazetteers for low-resource named entity recognition}.
\newblock In \emph{Proceedings of the 58th Annual Meeting of the Association
  for Computational Linguistics}, pages 8118--8123, Online. Association for
  Computational Linguistics.

\bibitem[{Samdani et~al.(2012)Samdani, Chang, and Roth}]{samdani2012unified}
Rajhans Samdani, Ming-Wei Chang, and Dan Roth. 2012.
\newblock Unified expectation maximization.
\newblock In \emph{NAACL-HLT}.

\bibitem[{Shen et~al.(2018)Shen, Lin, Huang, and Courville}]{shen2017neural}
Yikang Shen, Zhouhan Lin, Chin-Wei Huang, and Aaron Courville. 2018.
\newblock Neural language modeling by jointly learning syntax and lexicon.
\newblock In \emph{ICLR}.

\bibitem[{Shen et~al.(2019)Shen, Tan, Sordoni, and Courville}]{shen2018ordered}
Yikang Shen, Shawn Tan, Alessandro Sordoni, and Aaron Courville. 2019.
\newblock Ordered neurons: Integrating tree structures into recurrent neural
  networks.
\newblock In \emph{International Conference on Learning Representations
  (ICLR)}.

\bibitem[{Shi et~al.(2019)Shi, Mao, Gimpel, and Livescu}]{Shi_2019}
Haoyue Shi, Jiayuan Mao, Kevin Gimpel, and Karen Livescu. 2019.
\newblock Visually grounded neural syntax acquisition.
\newblock In \emph{Association for Computational Linguistics}.

\bibitem[{Shi et~al.(2021)Shi, {\.I}rsoy, Malioutov, and Lee}]{shi2021learning}
Tianze Shi, Ozan {\.I}rsoy, Igor Malioutov, and Lillian Lee. 2021.
\newblock \href {https://doi.org/10.18653/v1/2021.naacl-main.234} {Learning
  syntax from naturally-occurring bracketings}.
\newblock In \emph{NAACL-HLT}.

\bibitem[{Smith and Eisner(2005)}]{smith2005contrastive}
Noah~A Smith and Jason Eisner. 2005.
\newblock Contrastive estimation: Training log-linear models on unlabeled data.
\newblock In \emph{Proceedings of the 43rd Annual Meeting on Association for
  Computational Linguistics}.

\bibitem[{Snyder et~al.(2009)Snyder, Naseem, and
  Barzilay}]{snyder-etal-2009-unsupervised}
Benjamin Snyder, Tahira Naseem, and Regina Barzilay. 2009.
\newblock \href {https://www.aclweb.org/anthology/P09-1009} {Unsupervised
  multilingual grammar induction}.
\newblock In \emph{Proceedings of the Joint Conference of the 47th Annual
  Meeting of the {ACL} and the 4th International Joint Conference on Natural
  Language Processing of the {AFNLP}}, pages 73--81, Suntec, Singapore.
  Association for Computational Linguistics.

\bibitem[{S{\o}gaard(2017)}]{sogaard-2017-using}
Anders S{\o}gaard. 2017.
\newblock \href {https://aclanthology.org/W17-6310} {Using hyperlinks to
  improve multilingual partial parsers}.
\newblock In \emph{Proceedings of the 15th International Conference on Parsing
  Technologies}, pages 67--71, Pisa, Italy. Association for Computational
  Linguistics.

\bibitem[{Spitkovsky et~al.(2010)Spitkovsky, Jurafsky, and
  Alshawi}]{spitkovsky-etal-2010-profiting}
Valentin~I. Spitkovsky, Daniel Jurafsky, and Hiyan Alshawi. 2010.
\newblock \href {https://aclanthology.org/P10-1130} {Profiting from mark-up:
  Hyper-text annotations for guided parsing}.
\newblock In \emph{Proceedings of the 48th Annual Meeting of the Association
  for Computational Linguistics}, pages 1278--1287, Uppsala, Sweden.
  Association for Computational Linguistics.

\bibitem[{Stern et~al.(2017)Stern, Andreas, and
  Klein}]{stern-etal-2017-minimal}
Mitchell Stern, Jacob Andreas, and Dan Klein. 2017.
\newblock A minimal span-based neural constituency parser.
\newblock In \emph{Proceedings of the 55th Annual Meeting of the Association
  for Computational Linguistics (Volume 1: Long Papers)}.

\bibitem[{Swayamdipta et~al.(2018)Swayamdipta, Thomson, Lee, Zettlemoyer, Dyer,
  and Smith}]{Swayamdipta2018SyntacticSF}
Swabha Swayamdipta, Sam Thomson, Kenton Lee, Luke Zettlemoyer, Chris Dyer, and
  Noah~A. Smith. 2018.
\newblock \href {https://doi.org/10.18653/v1/D18-1412} {Syntactic scaffolds for
  semantic structures}.
\newblock In \emph{Proceedings of the 2018 Conference on Empirical Methods in
  Natural Language Processing}, pages 3772--3782, Brussels, Belgium.
  Association for Computational Linguistics.

\bibitem[{Taylor et~al.(2003)Taylor, Marcus, and Santorini}]{Taylor2003ThePT}
Ann Taylor, Mitchell Marcus, and Beatrice Santorini. 2003.
\newblock \href {https://doi.org/10.1007/978-94-010-0201-1_1} {The penn
  treebank: An overview}.
\newblock In \emph{Treebanks: Building and Using Parsed Corpora}, pages 5--22,
  Dordrecht. Springer Netherlands.

\bibitem[{Verspoor et~al.(2011)Verspoor, Cohen, Lanfranchi, Warner, Johnson,
  Roeder, Choi, Funk, Malenkiy, Eckert, Xue, Jr., Bada, Palmer, and
  Hunter}]{Verspoor2011ACO}
Karin Verspoor, Kevin Cohen, Arrick Lanfranchi, Colin Warner, Helen~L. Johnson,
  Christophe Roeder, Jinho~D. Choi, Christopher Funk, Yuriy Malenkiy, Miriam
  Eckert, Nianwen Xue, William A.~Baumgartner Jr., Michael Bada, Martha Palmer,
  and Lawrence~E. Hunter. 2011.
\newblock A corpus of full-text journal articles is a robust evaluation tool
  for revealing differences in performance of biomedical natural language
  processing tools.
\newblock \emph{BMC Bioinformatics}, 13:207 -- 207.

\bibitem[{Williams et~al.(2018)Williams, Drozdov, and
  Bowman}]{Williams2018DoLT}
Adina Williams, Andrew Drozdov, and Samuel~R. Bowman. 2018.
\newblock Do latent tree learning models identify meaningful structure in
  sentences?
\newblock \emph{Transactions of the Association for Computational Linguistics},
  6:253--267.

\bibitem[{Xu et~al.(2020)Xu, Li, Zhu, Zhang, and Zhou}]{xu-etal-2020-improving}
Dongqin Xu, Junhui Li, Muhua Zhu, Min Zhang, and Guodong Zhou. 2020.
\newblock \href {https://doi.org/10.18653/v1/2020.emnlp-main.196} {Improving
  {AMR} parsing with sequence-to-sequence pre-training}.
\newblock In \emph{EMNLP}.

\bibitem[{Younger(1967)}]{CKY2}
Daniel~H. Younger. 1967.
\newblock \href {https://doi.org/https://doi.org/10.1016/S0019-9958(67)80007-X}
  {Recognition and parsing of context-free languages in time n3}.
\newblock \emph{Information and Control}, 10(2):189--208.

\bibitem[{Yu(2012)}]{Yu2012Prior}
Chun-Nam Yu. 2012.
\newblock \href {https://proceedings.mlr.press/v22/yu12.html} {Transductive
  learning of structural {SVMs} via prior knowledge constraints}.
\newblock In \emph{Proceedings of the Fifteenth International Conference on
  Artificial Intelligence and Statistics}, volume~22 of \emph{Proceedings of
  Machine Learning Research}, pages 1367--1376, La Palma, Canary Islands. PMLR.

\bibitem[{Yu and Joachims(2009)}]{Yu2009LearningSS}
Chun-Nam~John Yu and Thorsten Joachims. 2009.
\newblock \href {https://doi.org/10.1145/1553374.1553523} {Learning structural
  {SVMs} with latent variables}.
\newblock In \emph{ICML}, pages 1169--1176.

\end{thebibliography}
\bibliographystyle{acl_natbib}

\newpage

\appendix

\clearpage
\section{Appendix}
\subsection{Constraint Statistics \label{sec:app_constraint}}

Here we report a detailed breakdown of span constraints and the associated constituent types. Compared with \citet{shi2021learning}, span constraints based on entities are less diverse with respect to constituent type. In future work, we plan to use their data combined with DIORA and PS-SVM training. Also, we hypothesize that RoBERTa would be effective as a data augmentation to easily find new constraints.

\begin{table}[ht!]
\setlength\tabcolsep{4pt}
\begin{center}
\resizebox{\columnwidth}{!}{%
\begin{tabular}{l | r r r |r  }
\toprule
& \multicolumn{3}{c|}{Ontonotes} & \multicolumn{1}{c}{CRAFT}  \\
& NER & Gazetter & PMI  & NER   \\
\midrule
Exact match & 96.3  & 51.3 & 43.9 & 57.4 \\
Conflict & 1.9  & 5.0 & 7.4  & 12.0 \\
\midrule
NP & 9.2 & 1.7 & 1.9 & 4.0 \\ 
VP & 0.0 & 0.0 & 0.0  & 0.0  \\
S & 0.1 & 0.0 & 0.0  & 0.0  \\ 
ADVP & 7.5 & 0.0 & 1.5   & 0.2  \\ 
ADJP & 3.1 & 0.8 & 2.2   & 3.3  \\
SBAR & 0.0 & 0.0 & 0.0   & 0.0  \\
NML & 21.6 & 11.6 & 14.9   & 17.9  \\
QP & 46.6 & 0.0 & 0.0   & 0.0  \\
PP & 0.1  &0.0  &0.0     & 0.0 \\
\textbf{Total} & 3.1 & 1.2 & 1.7   & 4.0  \\
\midrule
Number of sentences & \multicolumn{3}{c|}{115,811} & 18,951 \\
Number of ground truth spans & \multicolumn{3}{c|}{1,878,737} & 361,394 \\
Span/sentences & 0.50 & 0.19 & 0.28 & 0.77 \\
\bottomrule
\end{tabular}
}
\end{center}
\vspace{-0.4cm}
\caption{Statistics of different type constraints in Ontonotes. The top part shows how each constraint type agree with the ground truth parsing. The middle shows the percentages of each constituency spans found in constraint spans. The bottom part shows the total number of sentences and constraint spans per sentence.}
\label{tab:onto_stat}
\end{table}
% TODO: Make sure we reference the Lillian Lee work since this table is so similar.

\subsection{Hyperparameters \label{sec:app_hyperparam}}

We run a small grid search with multiple random seeds. The following search parameters are fixed for all experiments.

\begin{align*}
&\text{Model Dimension: } &400 \\
&\text{Optimization Algorithm: } &\text{Adam} \\
&\text{Hardware: } &\text{1x1080ti} \\
&\text{Training Time: } &O(\text{24h}) \\
\end{align*}

\noindent
Also, we search over the 4 variants of PS-SVM (\S\ref{sec:ps_variants}) when incorporating constraints. We mention the best performing variant of PS-SVM where it is relevant. The best performing setting for each hyperparameter is \underline{underlined}.

\newpage
\subsubsection{Newswire}

For newswire experiments, we train with Ontonotes and validate with PTB.

\begin{align*}
&\text{Learning Rate: } & \underline{2^{-3}},1^{-3} \\
&\text{Max Training Length: } &40 \\
&\text{Batch Size: } & 32 \\
&\text{Max Epochs: } &40 \\
&\text{Stopping Criteria: } &\text{Validation F1} \\
&\text{No. of Random Seeds: } &\text{6}
\end{align*}

\noindent
Using $\textsc{Rescale}$ gave the best result with ground truth entity-based constraints, and $\textsc{NCBL}$ gave the best result for PMI and gazetteer-based constraints.

\subsubsection{Biomedical Text}

First, to assist with domain adaptation, we train using a concatenation of CRAFT and MedMentions (DIORA$_{ft}$). We sample 3k sentences from CRAFT training data to use for validation.

\begin{align*}
&\text{Learning Rate: } &2^{-3} \\
&\text{Max Training Length: } &30 \\
&\text{Batch Size: } &32 \\
&\text{Max Epochs: } &20 \\
&\text{Stopping Criteria: } &\text{Validation F1} \\
&\text{No. of Random Seeds: } &\text{1}
\end{align*}

\noindent
Then we incorporate constraints and train only with CRAFT, using the same sample for validation.

\begin{align*}
&\text{Learning Rate: } & 2^{-3}, \underline{1^{-3}}, 5^{-4}, 1^{-4} \\
&\text{Max Training Length: } &40 \\
&\text{Batch Size: } &4,\underline{8},32 \\
&\text{Max Epochs: } &40 \\
&\text{Stopping Criteria: } &\text{Validation F1} \\
&\text{No. of Random Seeds: } &\text{3}
\end{align*}

\noindent
Using $\textsc{Structured Ramp}$ gave the best result.

\subsubsection{Other Details}

We report validation and test performance where applicable. All of our model output are shared in our github repo for further analysis. Training with PS-SVM uses the same parameters as standard DIORA training --- the supervision is directly on the scores computed for the inside-pass and does not require any new parameters.

% \subsubsection{Best}

% \begin{align*}
% &\text{Learning Rate: } &2^{-3} \\
% &\text{Max Training Length: } &40 \\
% &\text{Batch Size: } &32 \\
% &\text{Max Epochs: } &40 \\
% &\text{Stopping Criteria: } &\text{Validation F1} \\
% \end{align*}

% \subsubsection{\textsc{Rescale}}
% \begin{align*}
% &\text{Learning Rate: } &2^{-3} \\
% &\text{Max Training Length: } &40 \\
% &\text{Batch Size: } &32 \\
% &\text{Max Epochs: } &40 \\
% &\text{Stopping Criteria: } &\text{Validation F1} \\
% \end{align*}

% \subsubsection{\textsc{Min Difference}}
% \begin{align*}
% &\text{Learning Rate: } &1^{-3} \\
% &\text{Max Training Length: } &40 \\
% &\text{Batch Size: } &32 \\
% &\text{Max Epochs: } &40 \\
% &\text{Stopping Criteria: } &\text{Validation F1} \\
% \end{align*}

% \subsubsection{\textsc{Structured Ramp}}
% \begin{align*}
% &\text{Learning Rate: } &1^{-3} \\
% &\text{Max Training Length: } &40 \\
% &\text{Batch Size: } &32 \\
% &\text{Max Epochs: } &40 \\
% &\text{Stopping Criteria: } &\text{Validation F1} \\
% \end{align*}

\newpage

\subsubsection{Use of Validation Data}

\citet{Shi_2019} point out that validation sets can disproportionally skew performance of unsupervised parsing systems. We re-did early stopping using 100 random sentences and found that the best model remained the same in all cases. This is consistent with the DIORA-related experiments in \citet{Shi_2019} that show DIORA performance is robust when only a small number of samples are used for model selection.

\subsubsection{Why fine-tune?}

To be resource efficient, we use the pre-trained DIORA checkpoint from \citet{Diora} and fine-tune it for parsing biomedical text. DIORA was trained for 1M gradient updates on nearly 2M sentences from NLI data, taking 3 days using 4x GPUs. MedMentions has $\sim$40k training sentences, CRAFT has only $\sim$40k, and our PS-SVM experiments run in less than 1 day using a single GPU. 

% The improved results from training with only reconstruction on the MedMention + CRAFT data suggests that it may be possible to build an unsupervised parser solely from biomedical text, although we consider this out of the scope of our paper and leave this for future work. We will add any missing information (such as the number of sentences used in training) to make it more clear why we chose to fine-tune.

\subsection{Additional Related Work \label{sec:app_related_work}}

In the main text, we mention the most closely related work for training DIORA with our PS-SVM objective. Here we cover other work not discussed. Unsupervised parsing has a long and dense history, and we hope this section provides context to the state of the field, our contribution in this paper, and can serve as a guide for the interested researcher.

% \footnote{The related work in this paper and section has a focus on unsupervised phrase structure (i.e. constituency) parsing. There is also much work in unsupervised dependency parsing, although not necessarily mentioned here.}

\paragraph{History of unsupervised parsing over the last thirty years} As early as 1990, researcher were using corpus statistics to induce grammar, not unlike how our span constraints based on PMI are derived \cite{brill-etal-1990-deducing} --- at this point the Penn Treebank was still being built. Other techniques focused on optimizing sentence likelihood with probabilistic context-free grammars, although with limited success \cite{lari1990estimation,Carroll1992TwoEO,pereira-schabes-1992-inside}. Later work exploited the statistics between phrases and contexts \cite{Clark2001UnsupervisedIO,Klein2001NaturalLG}, but the most promising practical progress was not seen until 15+ years later.

In the mid 2010s, many papers were published about neural models for language that claimed to induce tree-like structure, albeit none made strong claims about unsupervised parsing. \citet{Williams2018DoLT} analyzed these models and discovered a negative result. Despite their tree-structured inductive bias, when measured against ground truth parse trees from the Penn Treebank these models did only slightly better than random and were not competitive with earlier work grammar induction. Shortly after, \citet{shen2017neural} developed a neural language model with a tree-structured attention pattern and \citet{htut-etal-2018-grammar} demonstrated its effectiveness at unsupervised parsing, the first positive result for a neural model. In quick succession, more papers were published with improve results and new neural architectures \cite[inter alia]{shen2018ordered,Diora,kim2019compound,kim-etal-2019-unsupervised,cao-etal-2020-unsupervised}, some of which we include as baselines in Table \ref{tab:parsing_entities}. Perhaps one of the more interesting work was improved performance of unsupervised parsing with PCFG when parameterized as a neural model (Neural PCFG; \citealp{kim2019compound}). These results suggest that the modern NLP machinery has made unsupervised parsing more viable, yet it is still not clear which of the newly ubiquitous tools (word vectors, contextual language models, adaptive optimizers, etc.) makes the biggest impact.

\paragraph{Variety of approaches to unsupervised parsing} The majority of the models in the work reported above optimize statistics with respect to the training data (with \citealp{cao-etal-2020-unsupervised} as an exception), but many techniques have been explored by now towards the same end. Unsupervised constituency parsing can be done in a variety ways including: exploiting patterns between images and text \cite{Shi_2019}, exploiting patterns in parallel text \cite{snyder-etal-2009-unsupervised}, joint induction of dependency and constituency \cite{klein-manning-2004-corpus}, iterative chunking \cite{ponvert-etal-2011-simple}, contrastive learning \cite{smith2005contrastive}, and more.

\paragraph{Other constraint types} We focus on span constraints, especially those from entities or derived from a lexicon, and encourage those spans to be included in the model's prediction. Prior knowledge of language can be useful in defining other types of structural constraints. For instance, in \citet{mayhew-etal-2019-named} the distribution of NER-related tokens helps improve performance for low-resource languages. More relevant, \citet{jin_depth_pcfg} present a PCFG with bounded recursion depth. \citet{pmlr-v119-niculae20a} present a flexible optimization framework for incorporating constraints such as bounded recursion depth and demonstrate strong results on synthetic data. Multiple works use web markup to improve syntactic parsing \cite{spitkovsky-etal-2010-profiting,sogaard-2017-using}.

\end{document}